\documentclass[graybox]{svmult}

\usepackage{type1cm}        % activate if the above 3 fonts are
\usepackage{makeidx}         % allows index generation
\usepackage{graphicx}        % standard LaTeX graphics tool
\usepackage{multicol}        % used for the two-column index
\usepackage[bottom]{footmisc}% places footnotes at page bottom

\usepackage[a4paper]{geometry}
\usepackage{newtxtext}       %
\usepackage{newtxmath}       % selects Times Roman as basic font
\usepackage{todonotes}
\usepackage{algorithm}
\usepackage{algpseudocode}
\usepackage[frozencache]{minted}
\setminted{
    linenos,
    frame=lines,
    style=colorful,
    fontsize=\small,
    breaklines,
}

\usepackage[inline]{enumitem}
\usepackage{mathtools}
\usepackage{amsfonts}
\usepackage{color}
\usepackage{amsmath}
\usepackage{commath}
\usepackage{todonotes}
\usepackage{hyperref}
\usepackage{nicefrac}
\usepackage{url}
\usepackage{xcolor}
\usepackage{booktabs}
\usepackage[inline]{enumitem}
\usepackage{inconsolata}
\usepackage{eucal}
\usepackage{bbold}
\usepackage{forest}
\usepackage{multirow}

\newcommand{\inner}[1]{\langle #1 \rangle}
\newcommand{\pars}[1]{\left( #1 \right)}
\newcommand{\bracks}[1]{\left[ #1 \right]}

\newcommand{\R}{\mathbb{R}}
\newcommand{\Z}{\mathbb{Z}}
\newcommand{\EE}{\mathbb{E}}

\newcommand{\cX}{\mathcal{X}}
\newcommand{\cU}{\mathcal{U}}

\newcommand{\cH}{\mathcal{H}}

\newcommand{\cO}{\mathcal{O}}

\newcommand{\bx}{\mathbf{x}}
\newcommand{\bu}{\mathbf{u}}
\newcommand{\by}{\mathbf{y}}

\newcommand{\bi}{\mathbf{i}}
\newcommand{\bj}{\mathbf{j}}

\newcommand{\rff}{\varphi}
\newcommand{\rffoned}{\rff^{\mathsf{1D}}}

\newcommand{\DPm}[1]{\Phi^{\mathsf{DP}}_{#1}}
\newcommand{\DPonem}[1]{\Phi^{\mathsf{DP\text{-}1D}}_{#1}}
\newcommand{\TRP}{\mathsf{TRP}}
\newcommand{\TRPm}[1]{\Phi^{\mathsf{TRP}}_{#1}}
\newcommand{\CS}{\mathsf{CS}}
\newcommand{\TS}{\mathsf{TS}}
\newcommand{\TSm}[1]{\Phi^{\mathsf{TS}}_{#1}}
\newcommand{\FFT}{\mathsf{FFT}}

\newcommand{\Sig}{\mathsf{Sig}}
\newcommand{\SigPDE}{\mathsf{SigPDE}}

\newcommand{\kernel}{\mathrm{k}}
\newcommand{\Kernel}{\mathrm{K}}
\newcommand{\sigkernel}{\kernel_{\mathrm{\Sig}}}
\newcommand{\sigkernelpde}{\kernel_{\mathrm{\SigPDE}}}
\newcommand{\sigkernelm}[1]{\kernel_{{\mathrm{\Sig}_{#1}}}}
\newcommand{\paths}{\cX_\text{paths}}
\newcommand{\seq}{\cX_{\text{seq}}}

\newcommand{\bigO}[1]{\mathcal{O}\left( #1 \right)}

\newcommand{\len}[1]{\mathsf{L}_{#1}}

\mathtoolsset{showonlyrefs=true}

\makeindex             % used for the subject index
\begin{document}

\title*{A User's Guide to \texttt{KSig}: GPU-Accelerated Computation of the Signature Kernel}
\author{Csaba Tóth\textsuperscript{1}, Danilo Jr Dela Cruz\textsuperscript{2}, Harald Oberhauser\textsuperscript{3}}

\authorrunning{Csaba Tóth, Danilo Jr Dela Cruz, Harald Oberhauser}

\institute{Mathematical Institute, University of Oxford, Andrew Wiles Building, Radcliffe Observatory Quarter, Woodstock Rd, Oxford OX2 6GG\\\textsuperscript{1}\texttt{\href{mailto:toth@maths.ox.ac.uk}{toth@maths.ox.ac.uk}}\\
\textsuperscript{2}\texttt{\href{mailto:delacruz@maths.ox.ac.uk}{delacruz@maths.ox.ac.uk}}\\
\textsuperscript{3}\texttt{\href{mailto:oberhauser@maths.ox.ac.uk}{oberhauser@maths.ox.ac.uk}}}
\maketitle

% \abstract*{The signature kernel is a positive definite kernel for sequential data.
%     It inherits theoretical guarantees from stochastic analysis, has efficient algorithms for computation, and shows strong empirical performance.
%     In this short survey paper for a forthcoming Springer handbook, we give an elementary introduction to the signature kernel and highlight these theoretical and computational properties.}

% \abstract{The signature kernel is a positive definite kernel for sequential data.
%     It inherits theoretical guarantees from stochastic analysis, has efficient algorithms for computation, and shows strong empirical performance.
%     In this short survey paper for a forthcoming Springer handbook, we give an elementary introduction to the signature kernel and highlight these theoretical and computational properties.}
% \vspace{-10pt}
\abstract{The signature kernel is a positive definite kernel for sequential and temporal data that has become increasingly popular in machine learning applications due to powerful theoretical guarantees, strong empirical performance, and recently introduced various scalable variations.
In this chapter, we give a short introduction to \texttt{KSig}, a \texttt{Scikit-Learn} compatible Python package that implements various GPU-accelerated algorithms for computing signature kernels, and performing downstream learning tasks.
We also introduce a new algorithm based on tensor sketches which gives strong performance compared to existing algorithms.
The package is available at \texttt{\href{https://github.com/tgcsaba/ksig}{https://github.com/tgcsaba/ksig}}.}

%%%%%%%%%%%%%%%%%%%%%%%%%%%%%%%%%%%%%%%%%%%%%%%%%%%%%%%%%%%%%%%%%%%%%%%%%%%%

\section{Introduction}
%The signature kernel has emerged as a powerful tool for sequential data modeling due to its roots in stochastic analysis and rich theoretical properties. While its versatility and theoretical guarantees make it an appealing choice, earlier implementations suffered from certain computational bottlenecks, especially when scaling to large or high-dimensional datasets. The recently introduced Random Fourier Signature Features (RFSF) map \cite{toth2023random} has managed to alleviate computational limitations, aiding in the scalability of signature kernels through efficient low-rank approximations via random feature methods. These developments leverage tensor-based dimensionality reduction techniques to approximate the signature kernel in linear time while retaining high accuracy.
This paper introduces \texttt{KSig}, a \texttt{Scikit-Learn} compatible Python library that implements various GPU-accelerated signature kernel algorithms; this includes exact algorithms as well as recent scalable variations such as Random Fourier Signature Features (RFSF) \cite{toth2023random}, to enable practical applications on modern machine learning problems involving large-scale time series datasets.

\textbf{The Signature Kernel.}
Let $\cX$ be the input data domain, which can be a topological space or a set, and
\begin{align}\label{eq:static_kernel}
    \kernel: \cX \times \cX \to \R,
\end{align}
be a positive definite, symmetric kernel on $\cX$. Throughout we refer to $\kernel$ as the \emph{static kernel}.
The kernel $\kernel$ is parametrized by some hyperparameters $\theta_\kernel \in \Theta_{\kernel}$ on $\cX$; e.g.~$\cX=\R^d$ with $\kernel$ being the RBF kernel and $\theta$ denoting the bandwidth, i.e.
\begin{align}
    \kernel(\bx, \by) =
    \exp\pars{\frac{\norm{\bx-\by}^2}{2\sigma^2}} \quad \text{for } \bx, \by \in \R^d
\end{align}
where $\theta_{\kernel} = \sigma \in \Theta_{\kernel} = \R_+$ is the bandwidth of the RBF kernel.

The signature kernel construction \cite{kiraly_kernels_2019} turns the kernel $\kernel$ on $\cX$ into a positive definite kernel $\sigkernel$ on the set of sequences of arbitrary length in the set $\cX$,
\begin{align}
    \seq \coloneqq \{\bx=(\bx_0,\ldots,\bx_L) : L \ge 0,\,\bx_i \in \cX\},
\end{align}
where we use $0$-based indexing for sequences, and we denote the ``effective'' length of a sequence $\bx = (\bx_0, \dots, \bx_L) \in \seq$ by $\len{\bx} = L$.
The resulting kernel
\begin{align}
\sigkernel : \seq \times \seq \to \R,
\end{align}
is called the \emph{signature kernel} and is again a positive definite kernel on $\seq$ that has strong theoretical guarantees from stochastic analysis and scalable algorithms.
It has become increasingly popular in machine learning tasks and statistics for sequential and temporal data.
We refer to \cite{lee2023signature} for a recent survey and theoretical background.

\textbf{Hyperparameters.}
The signature kernel $\sigkernel$ inherits the hyperparameter set $\Theta_{\kernel}$ of the original kernel $\kernel$, but has additional hyperparameters.
The full set of hyperparameters consists of:
\begin{itemize}
    \item \textbf{Static hyperparameters $\Theta_{\kernel}$.} These are simply the hyperparameter of static kernel $\kernel$, e.g.~the bandwidth in case of the RBF kernel.
    \item \textbf{Truncation level} $\mathbb{N}_\infty \coloneqq \mathbb{N}\cup \{\infty\}$. The number of iterated integrals to use. Informally, this is somewhat analogous to the degree in polynomial regression.
    \item \textbf{Preprocessing options} $\Theta_{\text{Preprocess}}$. Time series preprocessing techniques, such as resampling, adding lagged channels, adding a time coordinate, etc.
    \item \textbf{Algebraic structure} $\Theta_{\text{Algebra}}$. The signature kernel is an inner product of iterated integrals.
    There is a classic ``geometric'' choice motivated from stochastic analysis how these integrals are approximated, but an insight from the machine learning application is that non-geometric approximations can often improve the complexity with no deterioration in performance or even improvements.
    \item \textbf{Normalization} $\Theta_{\text{Normalization}}$. Normalization makes the signature kernel statistically robust and there are various possibilities.
    \end{itemize}
We denote the resulting hyperparameter set of $\sigkernel$ with
\[
\Theta_{\sigkernel} \coloneqq  \Theta_{\kernel} \times \mathbb{N}_\infty \times \Theta_{\text{Preprocess}} \times \Theta_{\text{Algebra}} \times \Theta_{\text{Normalization}}.
\]
We emphasize that this is the full hyperparameter set and advise readers who are new or less familiar to just focus on the first two, namely $\Theta_{\kernel}$ and truncation level $\mathbb{N}_\infty \coloneqq \mathbb{N}\bigcup\{\infty\}$.
Optimizing only these two sets of hyperparameters often already leads to good performance. We note that although doing a full grid-search for the bandwidth hyperparameter $\sigma = \theta_\kernel \in \Theta_\kernel$ is recommended when using e.g.~the RBF kernel \eqref{equation:RBF Kernel}, we find that sometimes the median heuristic \cite{garreau2017large} can give a reasonable baseline value when the optimization budget is limited, although there is no theory to support this in case of the signature kernel. Additionally, we find that the median heuristic can be useful for defining the grid for the bandwidth $\sigma$ by parameterizing $\sigma$ in terms of multiples of the value given by the median heuristic on a logarithmic grid as done in in \cite[Sec.~4]{toth2023random}.
% The median heuristic  serves as a good baseline choice for setting the bandwidth of the RBF kernel, but if there is available computational budget optimizing it over a grid often leads to improvements in performance.
% \todo[inline]{@Csaba: do you agree? Cs: I actually think it's a good default choice but obviously optimizing is best if there is computational budget for that}
% for hyperparameter choice since they are not theoretically justified in the case of the signature kernel and do in general not lead to good results.
For a hashed out example of complete hyperparameter optimization, we refer to the experimental work \cite{morrill2020generalised}, while we refer to \cite{toth2023random} for a discussion of the most relevant choices that are used to achieve state-of-the-art performance on popular time series classification tasks.

\textbf{\texttt{KSig}: GPU-Accelerated Signature Kernels in Python.}
\texttt{KSig} is a Python implementation of various algorithms to compute the signature kernel, and models for running downstream experiments.
% related quantitites such as MMD.
It uses GPU acceleration via \texttt{CuPy} \cite{cupy_learningsys2017} and is conformant to the usual \texttt{Scikit-Learn} \cite{pedregosa2011scikit} API and pipeline.

To demonstrate \texttt{KSig} we focus on the arguably most elementary task in kernel computation, namely to compute a Gram matrix, i.e.~the $(N\times N)$-matrix
\[
\left(\sigkernel(\bx^{(i)},\bx^{(j)})\right)_{i,j=1,\ldots,N} \quad \text{for} \quad \bx^{(1)},\cdots,\bx^{(N)} \in \seq
\]
Modern kernel learning provides essentially two routes to accomplish this: the first one is to use a kernel trick to efficiently evaluate the computation of each entry of the Gram matrix.
We refer to this classic approach as the \emph{dual approach} since the key is the kernel trick which reformulates computations via kernel duality in a RKHS and thereby avoids the direct computation and storage of a feature map.
A simple toy example of this dual approach in \texttt{KSig} is given by the following code.
\begin{minted}{python}
import numpy as np
import ksig
M = 5  # Truncation level.
k = ksig.static.kernels.RBFKernel()  # Static kernel.
k_sig = ksig.kernels.SignatureKernel(  # Signature kernel.
    M, static_kernel=k)
# Declaration of inputs.
X = ...  # Shape (N, L, d).
# Compute Gram matrix.
K_XX = k_sig(X)  # Shape (N, N).
\end{minted}
The second way to compute a Gram matrix is to construct a \emph{low-dimensional} embedding into a Euclidean space $\Phi:\seq \to \mathbb{R}^F$, such that the following holds
\[
\sigkernel(\bx,\by) \approx \langle \Phi(\bx), \Phi(\by) \rangle_{\R^F}
\]
where $\approx$ denotes an approximation that often only holds in probability.
That is, unlike in the dual approach (``the kernel trick''), a ``feature map'' is directly computed and then an inner product in $\mathbb{R}^F$ is taken.
A key insight in the kernel learning literature was that although a positive definite kernel can always be written as an inner product of infinite-dimensional features, excellent approximations of the kernel map can be found by using \emph{low-dimensional, random} feature maps $\Phi$. Arguably, the most popular way to construct these is via the celebrated Random Fourier Feature approach \cite{rahimi2007random}.
In the context of the signature kernel, a naive application of Bochner's theorem, that RFFs are based on, is not possible since the domain $\seq$ is not a linear space.
However, more sophisticated constructions can produce such low-dimensional, random features maps\footnote{We emphasize that the coordinates of $\Phi(x)$ are not an approximation to the usual signature coordinates that consist of iterated integrals.
The only connection is that their inner product is similar with high probability.}.
Next, we illustrate in \texttt{KSig}, how one can use the RFSF-TRP variation from \cite{toth2023random}, and compute features $\Phi(\bx)$; see Section \ref{sec:trp} for more details on this primal feature variation.
\begin{minted}{python}
import numpy as np
import ksig
# Declare static feature and projection.
M, D = 100  # Truncation, feature and proj size.
static_feat = ksig.static.features.RandomFourierFeatures(  # RFF.
    n_components=D)
proj = ksig.projections.TensorizedRandomProjection(  # TRP.
    n_components=D)
# Declare primal signature kernel.
rfsf_trp = ksig.kernels.SignatureFeatures(  # RFSF-TRP map.
    n_levels=M, static_features=static_feat, projection=proj)
# Declaration of inputs.
X = ...  # Shape (N, L, d)
# Fit kernel and compute Gram matrix.
rfsf_trp.fit(X)
K_XX = rfsf_trp(X)  # Gram matrix of shape (N, N).
\end{minted}
\begin{remark}\label{rem:primal}
Here, we used the RFSF-TRP \cite{toth2023random} kernel to approximate the $N \times N$-Gram matrix of the signature kernel. However, this incurs an $N^2$ complexity in the number of samples \(N\) due to taking inner products of feature maps. Fortunately, the primal approach has another advantage. In many downstream applications, the computation and storage of the $N\times N$-Gram matrix can be avoided by reformulating calculations via the ``primal features'' $\Phi(\bx^{(1)}),\ldots, \Phi(\bx^{(N)})\in \mathbb{R}^F$.
For example, inversion in Gaussian Process Regression can be reduced from \(N^3\) time and \(N^2\) memory complexity to $N$ in both time and space. This is significant when $N$ is large, allowing us to scale to datasets with millions of time series.
In \texttt{KSig}, we can compute features by calling the \texttt{transform} method of \texttt{SignatureFeatures} class:
\begin{minted}{python}
P_X = rfsf_trp.transform(X)  # Compute features of shape (N, MD+1)
# Under the hood, the call method computes features and takes inner products.
print(np.linalg.norm(K_XX - P_X @ P_X.T))  # We check the results match.
\end{minted}
We refer to \cite{toth2023random} for a more detailed discussion of these points and just remark that for large datasets the primal approach is usually more beneficial.
\end{remark}

\subsection*{Outline}
% \todo{@Csaba:do you agree? I have no strong opinion here}
% Section~\ref{sec:dualalgs} discusses different versions of the dual approach, and demonstrates empirically how an instance of truncated signature ke. In particular, Section~\ref{sec:dualalgs} briefly discusses the different algorithms and computational complexity for the dual approach, these are essentialy the recursive computations of \cite{kiraly_kernels_2019} and the PDE discretization of \cite{salvi_signature_2021-1}.
% Section \ref{sec:dualalgs} does the same for the primal approach, which includes the double low-rank algorithms of \cite{kiraly_kernels_2019} and the Random Fourier Signature Feature approach of \cite{toth2023random}.
% Section \ref{sec:numerical} contains some simple numerical examples and visualizations \todo{@Danilo: this should be your main focus; Csaba has some remarks/ideas on this}.
% Section \ref{sec:conclusion} gives a short summary and outlook.
% Appendix A overviews the interface, class and inheritcance structure of the \texttt{KSig} package.
% Appendix B contains additional details that we did not include in the main text \todo{@Csaba: eg. the weighted coefficients, you had cool normalization idea for the coefficients, count sketch, etc.}
% Appendix C contains additional code examples
% \todo{@Danilo: we'll skip the error analysis section}

\textbf{Section \ref{sec:dualalgs}.} We introduce the dual approach for signature kernels, which employs the kernel trick. Section \ref{sec:sigkern} introduces the terminology for signature kernels, and Section \ref{sec:truncsigkern} introduces recursive computations for truncated signature kernels following \cite{kiraly_kernels_2019}. Section \ref{sec:pdesigkern} explores PDE-based formulations for the signature kernel \cite{salvi_signature_2021-1}. We highlight their computational complexity and GPU acceleration capabilities. Section \ref{sec:dualimpl} demonstrates the declaration, configuration, and real-world deployment of these kernels using \texttt{KSig} on moderate datasets. As these methods are computationally demanding for large datasets and long sequences, exhibiting quadratic memory complexity in both number of samples and sequence length, we consider the primal approach to signature kernels in the section.

\textbf{Section \ref{sec:primalalgs}.} We focus on approximating the signature kernel with low-rank features to improve scalability while retaining performance. Section \ref{sec:rff} introduces Random Fourier Features \cite{rahimi2007random}, the main tool that we use to build randomized finite-dimensional approximations to signature kernels. Section \ref{sec:rfsf} recalls the construction of Random Fourier Signature Features from \cite{toth2023random}, which is our starting point. Sections \ref{sec:dp}, \ref{sec:trp}, and \ref{sec:ts} build on RFSF via various random projection techniques, and constructs scalable versions of RFSF with mildly different computational complexities. Finally, we give examples in Section \ref{sec:primalimpl} on the declaration, configuration, and deployment of primal signature features.

\textbf{Section \ref{sec:scalability}.} The previously introduced methods are compared in Section \ref{sec:scalability}, where we provide comparisons of computational trade-offs and visualizations of scalability to benchmark the performance of the various approaches in terms of
\begin{enumerate*}[label=(\arabic*)]
    \item sequence length vs.~resource usage, i.e.~computation time and memory footprint;
    \item number of random feature components vs.~approximation accuracy and resource usage.
    \item signature truncation level vs.~approximation accuracy and resource usage.
\end{enumerate*}

\textbf{Section \ref{sec:conclusion} and Appendix \ref{app:interface}.} We conclude with a brief summary and discussion of future directions, while the appendix overviews the \texttt{KSig} package interface.
% , while Appendices B and C contain additional details and example code. \todo[inline]{sync with what's actually there}

\section{Dual Signature Algorithms} \label{sec:dualalgs}
In this section, we detail variations of the signature kernel based on the dual approach -- that is, approaches which employ the kernel trick for computing Gram matrices.

\subsection{Signature Kernels} \label{sec:sigkern}
First, we establish some terminology.
Define the collection of all (non-strictly) ordered $m$-tuples
\begin{align}
    \Delta_m(L) = \{1 \leq i_1 \leq \cdots i_m \leq L \,:\, i_1, \dots, i_m \in \Z_+\},
\end{align}
where $m, L \in \Z_+$. Then, recall from \cite{kiraly_kernels_2019} that the order-$p$ and level-$m$ signature kernel for $p,m \in \Z_+$ between sequences $\bx, \by \in \seq$ can be written as
\begin{align} \label{eq:sigkernelm}
    \sigkernelm{m}(\bx, \by) = \sum_{\substack{\bi \in \Delta_m(\len{\bx}) \\ \bj \in \Delta_m(\len{\by})}} \frac{1}{\bi!^{(p)}} \nabla_{i_1, j_1} \kernel(\bx_{i_1}, \by_{j_1}) \cdots \nabla_{i_m, j_m} \kernel(\bx_{i_m}, \by_{j_m}),
\end{align}
where $\nabla_{i, j} \kernel(\bx_i, \by_j) = \kernel(\bx_i, \by_j) - \kernel(\bx_{i-1}, \by_j) - \kernel(\bx_i, \by_{j-1}, \bx_i) + \kernel(\bx_{i-1}, \by_{j-1})$ is a double difference operator, and $\bi!^{(p)}$ is such that if there are $k \in \Z_+$ distinct entries in the multi-index $\bi$, and respectively each is repeated $p_1, \dots, p_k$ times, then it is
\begin{align}
    \bi!^{(p)} = \begin{cases}
        p_1! \cdots p_k! \quad &\text{if } p_1, \dots, p_k \leq p \\
        0 & \text{otherwise}.
    \end{cases}
\end{align}

\subsection{Truncated Signature Kernels} \label{sec:truncsigkern}
Here, we discuss the dynamic programming based recursive algorithm proposed in \cite{kiraly_kernels_2019}. This approach computes exactly the signature kernel for finite truncation-$M$
\begin{align}\label{eq:ksig finite}
    \sigkernel(\bx, \by) = \sum_{m=0}^M \sigkernelm{m}(\bx, \by).
\end{align}
This is performed by an efficient recursion relation across levels. For a sequence $\bx \in \seq$, let $\bx_{0:k} = (\bx_0, \dots, \bx_k)$ denote a slicing operation, where $k \leq \len{\bx}$. Let $\kernel_{\bx_i} = \kernel(\bx_i, \cdot) \in \cH_\kernel$ denote the reproducing kernel lift of $\bx_i \in \cX$. Then, a consequence of the Chen identity for signatures \cite{lyons2007differential} is that the following recursion holds for the order-$p$ level-$m$ (reproducing) signature kernel \eqref{eq:sigkernelm} across time steps for $\bx \in \seq$
\begin{align}
    \sigkernelm{m}(\bx_{0:k}, \cdot) = \sigkernelm{m}(\bx_{0:k-1}, \cdot) + \sum_{q=1}^p \sigkernelm{m-q}(\bx_{0:k-1}, \cdot) \otimes \frac{(\nabla_k \kernel_{x_k})^{\otimes q}}{q!},
\end{align}
and unrolling this along the sequence with respect to the first term, we get
\begin{align}
    \sigkernelm{m}(\bx_{0:k}, \cdot) = \sum_{\kappa=1}^k \sum_{q=1}^p \sigkernelm{m-q}(\bx_{0:\kappa-1}, \cdot) \otimes \frac{(\nabla_\kappa \kernel_{\bx_\kappa})^{\otimes q}}{q!}.
\end{align}
Then, taking inner products and employing the reproducing property, we get that the order-$p$ level-$m$ signature kernel can be computed for $\bx, \by \in \seq$
{\small
\begin{align} \label{eq:sigkernelm_rec}
    \sigkernelm{m}&(\bx_{0:k}, \by_{0:l}) \\ =& \sum_{\kappa=1}^k \sum_{\lambda=1}^l \sum_{q=1}^p \sum_{r=1}^p \frac{1}{q! r!} \inner{\sigkernelm{m-q}(\bx_{0:\kappa-1}, \cdot) \otimes (\nabla_\kappa \kernel_{\bx_\kappa})^{\otimes q}, \sigkernelm{m-r}(\by_{0:\lambda-1}, \cdot) \otimes (\nabla_\lambda \kernel_{\bx_\lambda})^{\otimes r}}_{\cH_\kernel^{\otimes m}}.
\end{align}}
Although the computation of \eqref{eq:sigkernelm_rec} is rather involved, note that it only depends on the signature kernel levels lower than $m$, establishing a recursion across levels that allows to iteratively compute the signature kernel up to truncation level-$M$. The algorithm derived in \cite[App.~B]{kiraly_kernels_2019} is based on dynamic programming, and we refer to Algorithm 6 for the computation, which is also implemented in \texttt{KSig}. It has a complexity of $\cO((Mp^2+d) \len{\bx} \len{\by})$, where $M$ is the truncation level, $d$ is the complexity of evaluating $\kernel$, and $\len{\bx}, \len{\by}$ are respectively the effective length of the input sequences $\bx, \by$. We note this algorithm solely uses cumulative sum operations for the recursion step, which is completely parallelizable and using a work-efficient scan algorithm can be computed in logarithmic time given enough computing cores. Hence, the best achievable time complexity on GPUs actually scales in length as $\cO(\log \len{\bx} \log \len{\by})$, but the quadratic complexity remains in terms of memory, which can become a bottleneck.
Further, note this algorithm is exact, that is it computes \eqref{eq:ksig finite} without any approximation (which is not the case for many of the following algorithms).

Next, we discuss the role of the order parameter $p$. It allows to change the algebraic embedding employed by the signature kernel. We do not discuss the theory behind different choices of algebraic embeddings in detail, and instead refer to \cite{toth_seq2tens_2021,diehl2020time, diehl2023generalized}.
for discussion. However, we note that lowering $p$ leads to computational savings, while often retaining or even improving performance on downstream tasks. This can also be theoretically explained for the $p=1$ case by a universality result proved in \cite{toth_seq2tens_2021}.
Hence, it is often beneficial to start with $p=1$ for efficiency.

\subsection{Signature-PDE Kernels} \label{sec:pdesigkern}

An alternative approach for computing signature kernels proposed by \cite{salvi_signature_2021-1} is involved with approximating the untrucated signature kernel by a PDE-based approach -- that is, we set $M = \infty$, and we have
\begin{align}\label{eq:ksig untruncated}
    \sigkernelpde(\bx, \by) =\sum_{m=0}^\infty \sigkernelm{m}(\bx, \by),
\end{align}
where $\bx, \by \in \seq$.
Since the method is based on PDE discretization, we briefly consider the continuous-time formulation. Let $\bx, \by \in \paths$ be such that $(\kernel_{\bx_s})_{s \in [0, S]}$ and $(\kernel_{\by_t})_{t \in [0, T]}$ are bounded variation in $\cH_\kernel$. Let us define the function $\Kernel: [0, S] \times [0, T] \to \R$ as the signature kernel restricted to a given timeframe for $s \in [0, S]$ and $t \in [0, T]$
\begin{align}
    \Kernel(s, t) = \sigkernelpde(\bx\vert_{[0,s]}, \by\vert_{[0, t]}).
\end{align}
Then, they show in Section 2.2 that $\Kernel$ solves the following PDE
\begin{align}
\frac{\partial^2 \Kernel(s, t)}{\partial s \partial t} = \inner{\dot{\kernel_{\bx_s}}, \dot{\kernel_{\bx_t}}}_{\cH_k} \Kernel(s, t),
\end{align}
where $(\,\dot{}\,)$ denotes a time derivative. This is a Goursat problem, and by Taylor expansion they derive a 2\textsuperscript{nd}-order finite-difference scheme. Hence, we return to the discrete-time setting. Let $\bx, \by \in \seq$ be sequences corresponding to discretizations of underlying paths along the grid where the PDE is solved. Define
\begin{align}    \hat \Kernel(k, l) \approx \Kernel(k,l) \equiv \sigkernelpde(\bx_{0:k}, \by_{0:l})
\end{align}
as the discretized signature-PDE kernel over the restricted time frame up to steps $k$ and $l$. Then, by \cite[Def.~3.3]{salvi_signature_2021-1}, we can compute $\hat \Kernel$ recursively by
\begin{align} \label{eq:sigpde_rec}
    \hat \Kernel(k, l) &= \hat \Kernel(k, l-1) + \hat \Kernel(k-1, l) - \hat \Kernel(k-1, l-1) \\ &- \frac{1}{2} \nabla_{k, l} \kernel(\bx_k, \by_l) (\hat \Kernel(k, l-1) + \hat \Kernel(k-1, l)),
\end{align}
with the initial condition $\hat\Kernel(\cdot, 0) = \hat\Kernel(0, \cdot) = 1$, where $\nabla_{k,l}$ is the double difference operator as defined previously. Then, $\sigkernelpde(\bx, \by) = \hat\Kernel(\len{\bx}, \len{\by})$.
Note that unlike algorithm for the truncated signature kernel from
Section \ref{sec:truncsigkern}, this gives an approximation due to PDE discretization.
In practice, when starting from discrete-time data, the authors propose to apply dyadic refinement with linear interpolation in order to increase the accuracy of the PDE solver. We omit this here and leave it up to the user as a data preprocessing step to upsample the data.

In this algorithm, the recursion step requires computing the full matrix $(\hat \Kernel(k, l))_{k, l}$, which has a similar quadratic sequence length complexity as the dynamic programming algorithm above. In theory, complexity of computations is $\cO(d \len{\bx}\len{\by})$, where $d$ is the static kernel evaluation cost, and $\len{\bx}, \len{\by}$ are the lengths of the input time series. To speed up computations, \cite{salvi_signature_2021-1} proposes a parallelized recursion scheme, which opens the avenue for GPU based parallelization, by iterating over each antidiagonal of this matrix, and computing each entry of it in parallel. This potentially reduces the time complexity in $\len{\bx}$ and $\len{\by}$ to the number of antidiagonals, i.e.~$\cO(\len{\bx}+\len{\by})$, but it still retains the quadratic memory complexity as before.

In \texttt{KSig}, we provide an implementation using \texttt{Numba} \cite{lam2015numba} with a \texttt{CUDA} \texttt{JIT} kernel. We optimized the implementation compared to the original version
\footnote{The original implementation in \texttt{torch} is available at \href{https://github.com/crispitagorico/sigkernel}{https://github.com/crispitagorico/sigkernel}.}
released by the authors of \cite{salvi_signature_2021-1}. The main difference is that in there the whole $(\hat \Kernel(k, l))_{k, l}$ matrix is kept in memory during the computation, leading to a significant memory footprint, while our implementation only keeps the current and the previous two antidiagonals in memory saving considerable amounts of GPU memory. This allows for potential scalability to slightly longer time series, although other time series operations performed during the computation (i.e.~the static kernel evaluation) retain quadratic memory, hence, scalability to long time series is still a bottleneck.

\subsection{Implementation} \label{sec:dualimpl}

\textbf{Declaration.} We demonstrate how to instantiate $\sigkernel$ and $\sigkernelpde$ in \texttt{KSig}, detail their possible configurations, and finally show their real-world deployment.
\begin{minted}{python}
# Import packages.
import numpy as np
import ksig
# Declare kernel.
M, P = 5, 1  # Truncation and order.
bandwidth = 0.5  #  Bandwidth for RBF kernel.
k = ksig.static.kernels.RBFKernel(bandwidth)  # Set RBF as static kernel.
k_sig = ksig.kernels.SignatureKernel(  # Instantiate truncated SigKernel.
    n_levels=M, order=P, difference=True, normalize=True, static_kernel=k)
k_sig_pde = ksig.kernels.SignaturePDEKernel(  # Instantiate SigPDEkernel.
    difference=True, normalize=True, static_kernel=k)
# Declaration of inputs.
X = ...  # Shape (N, L, d).
# Compute Gram matrices.
K_XX_1 = k_sig(X)  # Shape (N, N).
K_XX_2 = K_sig_pde(X)  # Shape (N, N).
\end{minted}

\textbf{Configuration.} The details of the hyperparameter settings are as follows:
\begin{enumerate}[label=(\arabic*)]
    \item \texttt{static\_kernel} sets the static kernel to use, for which the choices are available in the package hierarchy under \texttt{ksig.static.kernels}. See Appendix \ref{app:static_kern}.
    \item \texttt{bandwidth} is required by any stationary kernel including the RBF;
    \item \texttt{n\_levels} sets the truncation level-$M$, exclusive to \texttt{SignatureKernel};
    \item \texttt{order} sets the algebraic embedding order-$p$, exclusive to \texttt{SignatureKernel};
    \item \texttt{difference} specifies whether to include the finite-differencing step, i.e. use sequence increments. Setting it to false corresponds to computing the signature of the cumulative sum of the lifted sequence as discussed under \cite[Rem.~4.10]{kiraly_kernels_2019};
    \item \texttt{normalize} specifies whether to normalize the kernel to unit norm, which can help with stability and consequently performance. For \texttt{SignatureKernel}, it individually normalizes signature levels. For $\texttt{SignaturePDEKernel}$, we do not have access to individual levels and normalizes the kernel globally.
\end{enumerate}

\textbf{Deployment.} Next, we demonstrate in a simple experiment how to use the constructed kernels for multivariate time series classification on a moderate-sized dataset from the UEA archive \cite{bagnall2018uea}. We will use the \texttt{TSLearn} library \cite{tavenard2020tslearn} for loading the dataset, and the \texttt{PrecomputedKernelSVC} model from \texttt{KSig} for support vector machine (SVM) classification \cite{steinwart2008support}, which provides a wrapper on top of the \texttt{Scikit-Learn} \cite{pedregosa2011scikit} SVM implementation with additional convenience features, such as cross-validation and minibatching. We consider the FingerMovements dataset, which contains electroencephaloGram(EEG) brain recordings represented as time series of $28$ channels and length $50$ with $316$ training examples and $100$ testing examples. The task is binary classification with roughly 50-50 distribution of classes.

\begin{minted}{python}
# Import packages and tools.
import numpy as np
import ksig
from ksig.models import PrecomputedKernelSVC
from sklearn.metrics.pairwise import euclidean_distances
from sklearn.preprocessing import LabelEncoder
from tslearn.datasets import UCR_UEA_datasets
# Load the dataset.
loader = UCR_UEA_datasets()
X_train, y_train, X_test, y_test = loader.load_dataset('FingerMovements')
# Rescale input time series.
scale = np.max(np.abs(X_train))
X_train /= scale
X_test /= scale
# Encode output labels.
encoder = LabelEncoder()
y_train = encoder.fit_transform(y_train)
y_test = encoder.transform(y_test)
# Use median heuristic for choosing bandwidth.
bandwidth = np.median(  # Need to flatten the time axis.
    euclidean_distances(X_train.reshape([X_train.shape[0], -1]))
) / np.sqrt(2.)
# Define signature kernel.
k = ksig.static.kernels.RBFKernel(bandwidth)  # Set RBF as static kernel.
k_sig = ksig.kernels.SignatureKernel(  # Instantiate truncated sigkernel.
    n_levels=5, order=1, difference=True, normalize=True, static_kernel=k)
# Define, fit and evaluate SVC model with cross-validation.
svc_grid = {'C': [1, 10, 100, 1000, 10000]}  # Define grid for SVC C.
model =  PrecomputedKernelSVC(k_sig, svc_grid=svc_grid)
model.fit(X_train, y_train)  # Train the model.
print(model.score(X_test, y_test))  # Achieves 0.6 accuracy.
\end{minted}

The previous code snippet demonstrates the full classification process from data loading, model training, and evaluation in a couple lines of code. We normalized the input dataset by its maximum absolute value, and encoded the output targets as binary labels. We employed the median heuristic \cite{garreau2017large} to set the bandwidth for the RBF kernel. Then, we defined the truncated signature kernel with order $p=1$, truncation $M=5$, and normalization. For the $C$ hyperparameter in SVM, we cross-validated over a grid of $C \in \{10^0, 10^1, \dots, 10^4\}$. After fitting and evaluating the model, we achieved an accuracy of $0.6$ on the testing set. In the benchmark study \cite{ruiz2021great}, the best reported result on this dataset is $0.56$, see Table 10, which we managed to outperform by minimal preprocessing and using essentially the default kernel hyperparameters. The average runtime of the computations above on an NVIDIA A100 GPU is 2.3 seconds and 430 megabytes of GPU memory. We note that by careful cross-validation over path preprocessing techniques and kernel hyperparameters, the performance of the signature kernel can be increased on this dataset to $0.64$, see \cite[Sec.~4]{toth2023random}.

\section{Primal Signature Algorithms}\label{sec:primalalgs}
Next, we discuss primal approaches -- that is, approaches which employ low-rank approximations to the Gram matrix of signature kernels allowing to construct finite-dimensional features such that their inner product approximate the signature kernel in probability. We focus on approximating the truncated signature kernel built from a continuous, bounded, and stationary static kernel on the sequence state space $\mathcal{X}=\mathbb{R}^d$, such as the the RBF kernel.

\subsection{Random Fourier Features} \label{sec:rff}
Our main tool will be the classic Random Fourier Features. \texttt{KSig} also provides other feature maps such as the Nyström method \cite{drineas2005nystrom}, but in practice there is not much difference in performance between the two, hence we restrict our attention to the former.

Recall that Random Fourier Features (RFF) \cite{rahimi2007random} are defined for a continuous, bounded, stationary kernel $\kernel: \R^d \times \R^d \to \R$ with spectral measure $\Lambda$ as
\begin{align} \label{eq:rff}
    \rff(\bx) = \frac{1}{\sqrt{D}}\pars{\cos(W^\top \bx), \sin(W^\top \by)} \quad \text{for } \bx \in \R^d,
\end{align}
where $\R^{d \times D} \ni W \sim \Lambda^D$. Another formulation of RFF is given by using
\begin{align} \label{eq:rff_1d}
    \rffoned(\bx) = \frac{1}{\sqrt{D}} \cos(W^\top \bx + \mathbf{b}),
\end{align}
where $\R^{d \times D} \ni W \sim \Lambda^D$ and $\R^D \ni \mathbf{b} \sim \cU^D(0, 2\pi)$, where $\cU$ refers to the uniform distribution. Then, both approaches offer an unbiased approximation of the kernel
\begin{align}
    \kernel(\bx, \by) = \EE[\inner{\rff(\bx), \rff(\by)}_{\R^{2D}}] = \EE[\inner{\rffoned(\bx), \rffoned(\by)}_{\R^D}]
\end{align}
with strong probabilistic concentration guarantees \cite{sriperumbudur2015optimal, szabo2019kernel, chamakh2020orlicz}. Note that both approaches use $D$ Monte Carlo samples for approximating the stationary static kernel $\kernel$, however, for RFF \eqref{eq:rff} each sample is $2$-dimensional leading to a $2D$-dimensional feature map, while for RFF-1D \eqref{eq:rff_1d} each sample is $1$-dimensional, leading to a $D$-dimensional feature map. This will be an important distinction later on, when discussing the diagonal projection approach in Section \ref{sec:dp}.

\subsection{Random Fourier Signature Features} \label{sec:rfsf}
First, we provide an overview of Random Fourier Signature Features (RFSF) from \cite{toth2023random}.
Let $\rff_1, \dots, \rff_m: \R^d \to \R^D$ be independent RFF maps as in \eqref{eq:rff}.
Then, the order-$p$ level-$m$ Random Fourier Signature Feature map is defined as
\begin{align} \label{eq:rfsf}
    \Phi_m(\bx) \coloneqq \sum_{\bi \in \Delta_m(\len{\bx})} \frac{1}{\bi!^{(p)}} \nabla_{i_1} \rff_1(\bx_{i_1}) \otimes \cdots \otimes \nabla_{i_m} \rff_m(\bx_{i_m}).
\end{align}
We leave it a straightforward exercise for the reader to check that this indeed provides an unbiased approximation to the level-$m$ order-$p$ signature kernel \eqref{eq:sigkernelm}.
Alternatively, we could have used in place of $\rff$ the lower-dimensional $\rffoned$, but for didactic purposes we shall focus on the former and later discuss the differences. The recursion to compute \eqref{eq:rfsf} can be written for $\bx \in \seq$ as
\begin{align} \label{eq:rfsf_rec}
    \Phi_m(\bx_{0:k}) = \sum_{\kappa=1}^k \sum_{q=1}^p \frac{1}{q!} \Phi_{m-q}(\bx_{0:\kappa-1}) \otimes \nabla_{\kappa} \rff_{m-q+1}(\bx_\kappa) \otimes \cdots \otimes \nabla_\kappa\rff_m(\bx_\kappa)
\end{align}
This feature map represents the order-$p$ level-$m$ signature kernel in feature space using $(2D)^m$ coordinates.
Alternatively, if we used $\rffoned$ in place of $\rff$ we would end up with a feature map that is $D^m$ dimensional, and we denote this variation by
\begin{align} \label{eq:rfsf_1d}
    \Phi_m^{\mathsf{1D}}: \seq \to \R^{D^m}.
\end{align}
An algorithm to compute \eqref{eq:rfsf_rec} was proposed by \cite[Alg.~D.1]{toth2023random}, which computes \eqref{eq:rfsf} with a complexity of $\cO(L(MdD + p D^m))$, the algorithm and complexity of $\varphi_m^\mathsf{1D}$ is analogous since the dimension of the static feature map is only different by a factor of $2$. Although this allows to approximate the signature kernel using a finite-dimensional space, it still suffers from the combinatorial explosion of signature coordinates incurred by high-degree tensors. This motivates the use of (randomized) dimensionality reduction techniques to aid in further scalability. We consider three projection techniques: \begin{enumerate*}[label=(\arabic*)] \item the diagonal projection and \item tensor random projection approaches proposed by \cite{toth2023random, toth2024learning}; \item a novel approach based on the tensor sketch \cite{pham2013fast} not discussed in previous work yet \end{enumerate*}.

\subsection{Diagonal Projection Approach} \label{sec:dp}
The first approach we consider is the diagonal projection approach of \cite{toth2023random}. Here, the core idea is that since RFSF in \eqref{eq:rfsf} is an unbiased estimator for the truncated signature kernel \eqref{eq:sigkernelm}, we can sample multiple independent copies of it and average them to get an estimator for the signature kernel. Then, by the law of large numbers and probabilistic concentration arguments, this estimator converges to the signature kernel $(1/m)$-subexponentially fast (see \cite[Thm.~3.5]{toth2023random}  as we increase the number of samples. In order to make the computation of each RFSF sample fast, for each of them we decrease the internal RFF size to $1$-sample. Thus, let $\Phi^{(1)}, \dots, \Phi^{(D)}: \seq \to \R^{2^m}$ be independent RFSF maps as in \eqref{eq:rfsf} with internal RFF sample size-$1$. Then, the diagonally projected RFSF map (RFSF-DP) is defined for $\bx \in \seq$ as
\begin{align} \label{eq:dp}
    \DPm{m}(\bx) = \frac{1}{\sqrt{D}} \pars{\Phi_m^{(1)}(\bx), \dots, \Phi_m^{(D)}(\bx)} \in \R^{2^m D},
\end{align}
where the computation of each $\Phi_m^{(i)}$ is completely analogous to \eqref{eq:rfsf_rec}, except that now each of these have internal RFF sample size-$1$, resulting in $2$-dimensional RFFs as per \eqref{eq:rff}, which leads to each $\Phi_m^{(i)}$ being $2^m$ dimensional, leading to an overall dimensionality of $2^m D$. See \cite[Alg.~D.2]{toth2023random} for the vectorized computation of this approach. The complexity of computing all features up to truncation-$M$ is then governed by computing \eqref{eq:rfsf} $D$-times in parallel with RFF size-$1$, which is $\cO(L(MdD + p D 2^{M+1}))$, which has an exponential factor in $M$, although with a base of $2$, which is mild for moderate truncation.

Next, we discuss the diagonal projection approach applied to the RFSF-1D variation \eqref{eq:rfsf_1d} \cite{toth2024learning}, that uses $1$-dimensional RFF samples from \eqref{eq:rff_1d}. We denote this approach by $\DPonem{m}: \seq \to \R^D$, and similarly to \eqref{eq:dp}, it is given by concatenating $D$ independent copies $\Phi_m^{\mathsf{1D}, (1)}, \dots, \Phi_m^{\mathsf{1D}, (D)}: \seq \to \R$, so we have
\begin{align} \label{eq:dp1d}
    \DPonem{m} = \frac{1}{\sqrt{D}} \pars{\Phi^{\mathsf{1D}, (1)}_m, \dots, \Phi^{\mathsf{1D}, (D)}_m} \in \R^D.
\end{align}
The algorithm to compute \eqref{eq:dp1d} is analogous to the computation of \eqref{eq:dp}, i.e.~computing the $D$ independent copies of $\Phi^{\mathsf{1D}, (i)}$ in parallel, and returning their concatenation as the resulting feature map. We call this variation RFSF-DP-1D that has a complexity of $\cO(LMD(d + p))$, hence avoiding the exponential factor $2^M$, resulting in better scalability in $M$. In practice, for given of $D$ and $M$ it is lower dimensional, but has higher variance due to the static feature map \eqref{eq:rff_1d} having higher variance than \eqref{eq:rff}.

\subsection{Tensor Random Projection Approach} \label{sec:trp}
This approach was also introduced by \cite{toth2023random}, and it uses tensor random projections (TRP) \cite{rakhshan2020tensorized, sun2021tensor} to reduce the dimensionality of the base RFSF map \eqref{eq:rfsf} by constructing factorized projection tensors using the tensor product of Gaussian vectors.

Let $X \in (\R^D)^{\otimes m}$. The rank-$1$ TRP of $X$ is constructed via the random functional $\ell \in L((\R^D)^{\otimes m}, \R)$, $\ell(X) = \inner{\bu_1 \otimes \cdots \otimes \bu_m, X}_{(\R^D)^{\otimes m}}$,
where $\bu_1, \dots, \bu_m \in \R^D$ have i.i.d. $\mathcal{N}(0, 1)$ entries. Then, stacking $Q$ independent functionals $\ell_1 = \inner{\bu_1^{(1)} \otimes \cdots \otimes \bu_m^{(1)}, \cdot}, \dots, \ell_Q = \inner{\bu_1^{(Q)} \otimes \cdots \otimes \bu_m^{(Q)}, \cdot}$ gives the TRP map
\begin{align} \label{eq:trp_def}
    \mathsf{TRP}(X) = \frac{1}{\sqrt{Q}} (\ell_1(X), \dots, \ell_Q(X))^\top \in \R^Q, \quad \TRP: (\R^D)^{\otimes m} \to \R^Q
    .
\end{align}
We can apply this idea to RFSF by composing it with the TRP map in order to reduce its dimensionality, such that $\TRPm{m}(\bx) = \TRP\pars{\Phi_m(\bx)}$ for $\bx \in \seq$. By exploiting linearity and that the inner product with the projection tensors factorizes across the components in the tensor product, we get the explicit formulation
\begin{align} \label{eq:trp}
    \TRPm{m}(\bx) = \frac{1}{\sqrt{Q}} \sum_{\bi \in \Delta_m(\len{\bx})} \frac{1}{\bi!^{(p)}} P_1(\nabla_{i_1} \rff_1(\bx_{i_1})) \odot \cdots \odot P_m(\nabla_{i_m} \rff_m(\bx_{i_m})),
\end{align}
where $P_j(\bx) = \pars{\inner{\bu_j^{(1)}, \bx}, \dots, \inner{\bu_j^{(Q)}, \bx}}^\top \in \R^Q$ for $\bx \in \R^D$. The recursive formulation in this case is written similarly to \eqref{eq:rfsf_rec} for a $\bx \in \seq$ as
\begin{align} \label{eq:trp_rec}
    \TRPm{m}&(\bx_{0:k}) \\ =& \sum_{\kappa=1}^k \sum_{q=1}^p \frac{1}{q!} \TRPm{m-q}(\bx_{0:k-1}) \odot P_{m-q+1}(\nabla_{k} \rff_{m-q+1}(\bx_k)) \odot \cdots \odot P_{m}(\nabla_{k} \rff_{m}(\bx_k)).
\end{align}
An algorithm to compute \eqref{eq:trp_rec} is given under \cite[Alg.~D.3]{toth2023random}. It has a complexity of $\cO(MLD(d + p Q))$. Typically \(Q = D\), for which it is more expensive than RFSF-DP-1D by a factor of $D$ due to performing multiplications with the $Q \times D$ projection matrices in \eqref{eq:trp_rec}.

\subsection{Tensor Sketch Approach} \label{sec:ts}
Finally, the last approach we consider is similar to the tensor random projection approach, where individual random projections are combined to build a projection of the tensor product. It is termed the tensor sketch introduced in \cite{pham2013fast}, and it starts with the count sketches \cite{charikar2002finding} of the component vector across the tensor product, and then combines the count sketches of these components to build a count sketch for the tensor product. The core idea is to use that the circular convolution of count sketches is again a count sketch corresponding to their tensor product. Further, it admits efficient computations by realizing convolutions in the frequency domain as element-wise products. Using this property, we can build the count sketch of RFSFs by replacing each tensor product in \eqref{eq:rfsf} with the convolution of count sketches.

We introduce the count sketch, and show how given count sketches of vectors we can build a count sketch for their tensor product, called the tensor sketch. Let $h: [d] \to [Q]$ and $s: [d] \to \{\pm 1\}$ be $2$-wise independent hash functions. Then, the count sketch $\CS: \R^d \to \R^Q$ of a $\bx = (x_1, \dots, x_d) \in \R^d$ is
\begin{align} \label{eq:cs}
    \CS(\bx) = \pars{\CS(\bx)_1, \dots \CS(\bx)_Q}^\top \in \R^Q, \quad \CS(\bx)_j = \sum_{h(i) = j} s(i) x_i.
\end{align}
Then an unbiasedness approximation result holds for $\bx, \by \in \R^d$ such that
\begin{align}
    \inner{\bx, \by}_{\R^d} = \EE\bracks{\inner{\CS(\bx), \CS(\by)}_{\R^Q}} \approx \inner{\CS(\bx), \CS(\by)}_{\R^Q},
\end{align}
with the variance of the estimator on the RHS is bounded by the $2$-norm of the inputs \cite[Lem.~3]{pham2013fast}.
Now, let $\bx_1, \dots, \bx_m \in \R^D$, and $\CS_1, \dots, \CS_m: \R^D \to \R^Q$ independent CS operators. Then, the tensor sketch operator $\TS: (\R^D)^{\otimes m} \to \R^Q$ is
\begin{align} \label{eq:ts_def}
    \TS(\bx_1 \otimes \cdots \otimes \bx_m) = \CS_1(\bx_1) \star \cdots \star \CS_m(\bx_m) \in \R^Q,
\end{align}
where $\star$ denotes circular convolution, and can be computed efficiently using the Fast Fourier Transform (FFT) in $\cO(Q \log Q)$ time. Note that TS is again a count sketch of $\bx_1 \otimes \cdots \otimes \bx_m \in (\R^D)^{\otimes m}$ \cite{pham2013fast}. Hence, we can build the TS of \eqref{eq:rfsf} for $\bx \in \seq$ by
\begin{align} \label{eq:ts}
    \TSm{m}(\bx) = \sum_{\bi \in \Delta_m(\len{\bx})} \frac{1}{\bi!^{(p)}} \CS_1(\nabla_{i_1} \rff_1(\bx_{i_1})) \star \cdots \star \CS_m(\nabla_{i_m} \rff_m(\bx_{i_m})).
\end{align}
Then, we establish a recursion analogous to \eqref{eq:rfsf_rec} for $\bx \in \seq$ given by
{\small
\begin{align} \label{eq:ts_rec}
    \TSm{m}&(\bx_{0:k}) \\
    =& \sum_{\kappa=1}^k \sum_{q=1}^p \frac{1}{q!} \TSm{m-q}(\bx_{0:\kappa-1}) \star \CS_{m-q+1}(\nabla_{\kappa} \rff_{m-q+1}(\bx_\kappa)) \star \cdots \star \CS_m(\nabla_\kappa\rff_m(\bx_\kappa)) \\
    =& \sum_{\kappa=1}^k \sum_{q=1}^p \frac{1}{q!} \FFT^{-1}\pars{\FFT(\TSm{m-q}(\bx_{0:\kappa-1})) \odot \FFT(\CS_{m-q+1}(\nabla_{\kappa} \rff_{m-q+1}(\bx_\kappa))) \odot \cdots \odot \FFT(\CS_m(\nabla_\kappa\rff_m(\bx_\kappa)))},
\end{align}}
where the second line follows from being able to realize circular convolutions $\star$ as element-wise products $\odot$ in the frequency domain. The algorithm to compute \eqref{eq:ts_rec} is similar to the one for TRP with appropriate changes, that is implemented in \texttt{KSig} under the same API. It has a time complexity of $\cO(ML(Dd + p Q \log Q)$. Setting \(Q = D\), it is more expensive than RFSF-DP-1D by a factor of $\log D$ due to the FFT computation, but cheaper than RFSF-TRP since it replaces the $D^2$ factor by a factor of $D \log D$.

\subsection{Implementation} \label{sec:primalimpl}
\textbf{Declaration.} We demonstrate the declaration, configuration and real-world deployment of each primal variation in \texttt{KSig} with \(D = Q\). A discussion on the choice of \(D, Q\) can be found in Section \ref{sec:scalability}.
\begin{minted}{python}
import numpy as np
import ksig
# Declare static features and projections.
bandwidth = 0.5  # Bandwidth for RFF.
n_components = 100  # Num static features and projections.
rff = ksig.static.features.RandomFourierFeatures(  # RFF.
    bandwidth, n_components=D)
rff_1d = ksig.static.features.RandomFourierFeatures1D(  # RFF-1D.
    bandwidth, n_components=D)
dp = ksig.projections.DiagonalProjection(n_components=D)  # DP.
dp_1d = ksig.projections.DiagonalProjection(  # DP-1D.
    n_components=D, internal_size=1)
trp = ksig.projections.TensorizedRandomProjection(  # TRP.
    n_components=D)
ts = ksig.projections.TensorSketch(  # TS.
    n_components=D)
# Declare primal signature kernels.
M, P = 5, 1  # Truncation, order.
rfsf_dp = ksig.kernels.SignatureFeatures(  # RFSF-DP.
    n_levels=M, order=P, static_features=rff, projection=dp,
    difference=True, normalize=True)
rfsf_dp_1d = ksig.kernels.SignatureFeatures(  # RFSF-DP-1D.
    n_levels=M, order=P, static_features=rff_1d, projection=dp_1d,
    difference=True, normalize=True)
rfsf_trp = ksig.kernels.SignatureFeatures(  # RFSF-TRP.
    n_levels=M, order=P, static_features=rff, projection=trp,
    difference=True, normalize=True)
rfsf_ts = ksig.kernels.SignatureFeatures(  # RFSF-TS.
    n_levels=M, order=P, static_features=rff, projection=ts,
    difference=True, normalize=True)
# Declaration of inputs.
X = ...  # Shape (N, L, d),
# Fit each kernel to the data.
rfsf_dp.fit(X)
rfsf_dp_1d.fit(X)
rfsf_trp.fit(X)
rfsf_ts.fit(X)
# Compute features using the transform method.
P_X_1 = rfsf_dp.transform(X)  # (N, D^(2^(M+1)-1)+1)
P_X_2 = rfsf_dp_1d.transform(X)  # (N, MD+1)
P_X_3 = rfsf_trp.transform(X)  # (N, MD+1)
P_X_4 = rfsf_ts.transform(X)  # (N, MD+1)
\end{minted}

\textbf{Configuration.} Note that each variation uses the same API of \texttt{SignatureFeatures}, which can be configured using the following settings:
\begin{enumerate}[label=(\arabic*)]
    \item \texttt{bandwidth} is required by RFF and RFF-1D, and analogous to the bandwidth of RBF;
    \item \texttt{n\_components} sets the sizes of static features, and for TRP and TS the projection;
    \item \texttt{n\_levels} sets the truncation level-$M$;
    \item \texttt{order} sets the algebraic embedding order-$p$;
    \item \texttt{difference} specifies whether to include the finite-differencing step;
    \item \texttt{normalize} specifies whether we should normalize the features to unit norm by individually normalizing each signature level;
    \item \texttt{static\_features}, the choice of the static feature to use, e.g. RFF, RFF-1D;
    \item \texttt{projection}, which determines the type of (random) projection technique used.
\end{enumerate}

\textbf{Deployment.} Finally, we experimentally demonstrate how low-rank signature kernels can be employed combined with a linear SVM implementation and using GPU acceleration for both kernel and downstream computations. \texttt{KSig} implements a wrapper on top of a GPU-accelerated linear SVM implementation from CUML \cite{raschka2020machine}, which can be instantiated as \texttt{ksig.models.PrecomputedFeatureLinSVC}, that similarly to before adds further convenience features, such as cross-validation, and minibatching for the feature computations in order to conserve GPU memory.

Next, we load the PenDigits dataset from the UEA repository, which is a larger handwritten digit classification task of time series with $2$ channels, length $8$, $7494$ training examples, $3498$ testing examples, and $10$ balanced classes.
\begin{minted}{python}
# Import packages and tools.
import numpy as np
import ksig
from ksig.models import PrecomputedFeatureLinSVC
from sklearn.metrics.pairwise import euclidean_distances
from sklearn.preprocessing import LabelEncoder
from tslearn.datasets import UCR_UEA_datasets
# Load the dataset.
loader = UCR_UEA_datasets()
X_train, y_train, X_test, y_test = loader.load_dataset("PenDigits")
# Rescale input time series.
scale = np.max(np.abs(X_train))
X_train /= scale
X_test /= scale
# Encode output labels.
encoder = LabelEncoder()
y_train = encoder.fit_transform(y_train)
y_test = encoder.transform(y_test)
# Use median heuristic for choosing bandwidth.
bandwidth = np.median(  # Need to flatten the time axis.
    euclidean_distances(X_train.reshape([X_train.shape[0], -1]))
) / np.sqrt(2.)
# Declare primal signature kernel.
D = 200  # RFF and projection size.
rff = ksig.static.features.RandomFourierFeatures(  # RFF.
    bandwidth, n_components=D)
trp = ksig.projections.TensorizedRandomProjection(n_components=D) # TRP.
rfsf_trp = ksig.kernels.SignatureFeatures(  #RFSF-TRP.
    n_levels=5, order=1, static_features=rff, projection=trp)
# Define, fit, and evaluate LinearSVC model with cross-validation.
svc_grid = {'C': [1, 10, 100, 1000, 10000]}  # Grid for SVC hyperparameter.
model =  PrecomputedFeatureLinSVC(  # LinearSVC model with batching and CV.
    rfsf_trp, svc_grid=svc_grid, need_kernel_fit=True, on_gpu=True, batch_size=1000)
model.fit(X_train, y_train)  # Fit model.
print(model.score(X_test, y_test))  # Achieves 0.961 accuracy.
\end{minted}
The model above performs similar data loading, and preprocessing steps as before, and uses the RFSF-TRP formulation of primal signature kernels with $D = 200$ components, $M=5$ truncation, and $p=1$ order, which leads to a $1001$-dimensional feature map. The RFF bandwidth is chosen by the median heuristic as before, and we do not tune it. The SVC hyperparameter is cross-validated over $C \in \{10^0, 10^1, 10^2, 10^3, 10^4\}$. Additionally, we set the batch size for the model to $1000$, which allows to perform all computations with a minimal GPU memory footprint. The average runtime of the computation is $4.6$ seconds and only uses around $600$ megabytes of GPU memory. We achieve an average testing accuracy of $0.961$ out of $5$ evaluations. Note that by careful fine-tuning of path preprocessing approaches and kernel hyperparameters, this can be increased to $0.98$, see \cite[Sec.~4]{toth2023random}.

\section{Scalability Analysis} \label{sec:scalability}
% \todo[inline]{@Danilo: Please do this section. Just make a lot of interesting plots and add some discussion regarding which variation is better in what regimes. Some ideas: 1) for all approaches sequence length vs computation time, and sequence length vs memory footprint; 2) for random feature approaches: number of components ($D$) vs computation time, and number of components vs GPU memory footprint, 3) number of components vs approximation accuracy (I think it's best measured in terms of $\EE\vert(\kernel - \tilde\kernel) / \kernel\vert$, i.e. the absolute deviation percetange, 4+5) for fixed number of components signature level vs approximation error and resource usage (time and memory).}

We study the memory footprint, runtime and accuracy of the algorithms in computing the Gram matrix of a synthetic dataset and observe the following:
\begin{itemize}
    \item \textbf{Primal methods are necessary for long sequences.} Running each algorithm until the memory limit, dual methods can process sequence lengths \(L \sim 10^3\) whereas primal methods can process \(L \ge 10^5\) (Fig. \ref{fig:Sequence Length Resources}). This is due to their linear dependence on the sequence length \(L\) (Table \ref{table:GramMatrix}).
    \item \textbf{Diagonal Projection methods achieve the highest accuracy.} For given memory constraint, we can increase the parameters of diagonal projection methods further yielding higher accuracy (Fig. \ref{fig:Feature Map Accuracy}).
    \item \textbf{RFSF-TRP and RFSF-TS generate better features.} For a given feature map size, the resulting feature maps yield higher accuracy (Fig. \ref{fig:Feature Map Accuracy}). For downstream applications, where the feature map size have a larger effect on the overall complexity, these methods may be more desirable.
    \item \textbf{Projection size is the accuracy bottleneck for RFSF-TRP and RFSF-TS.} A large static feature size \(D\) is redundant if the projection size \(Q\) is not sufficiently large (Fig. \ref{fig:Projection RFF MAPE}).
    \item \textbf{All algorithms are effectively linear in truncation level for fixed order.} With the feature map size equal across algorithms, we observe that each one scales linearly in memory and runtime (Fig.~\ref{fig:Truncation Level}).
\end{itemize}

Experiments were conducted on a NVIDIA A100 GPU with a memory capacity of 40GB. The synthetic dataset consists of \(d\)-dimensional Brownian Motion over \([0, 1]\). Across all experiments we fix dimension \(d = 5\) and order \(p = 1\). Other parameter choices vary and will be specified in the associated caption.

\subsection{Complexity}
To guide our intuition, we begin with a big \(\bigO{\cdot}\) complexity analysis of the algorithms restricted to the case of embedding order \(p=1\). Denote \(N\) for the number of sequences, \(L\) sequence length, \(d\) sequence dimension, \(M\) truncation level and \(D, Q\) for the \texttt{n\_components} parameter of \texttt{StaticFeatures} and \texttt{RandomProjection} respectively. We assume the partial ordering to obtain simpler expressions
\begin{equation}
    1 < M \le d \le Q, D  \le N, L
    .
\end{equation}
Furthermore, we assume the evaluation cost of the static kernel \(\kernel(\bx, \by)\) is \(\bigO{d}\).

The complexity to compute the Gram matrix is given in Table \ref{table:GramMatrix}. Dual methods \texttt{KSig} and \texttt{KSigPDE} are quadratic in sequence length \(L\) whereas primal methods are linear in sequence length \(L\).

\begin{table}[!h]
    \centering
    \begin{tabular}{l|l|l}
    \textbf{Name}    & \textbf{Time}         & \textbf{Space}   \\
    \hline
    \textbf{Primal} & \(N^2F + NLT\) & \(NF + NLS\) \\
    \textbf{KSig}      & \(N^2 L^2 M d\)        & \(N^2 L^2\)           \\
    \textbf{KSigPDE}     & \(N^2 L^2 d\) & \(N^2 L^2\)
    \end{tabular}
    \caption{Complexity to compute Gram matrices for dual methods and corresponding low-rank factors for primal methods. For primal methods, refer to Table \ref{table:Primal Methods Complexity} for concrete values of the variables \(F, T, S\), which are such that the \textit{Output} is \(N \times F\), \textit{Time} is \(NLT\) and \textit{Space} is \(NLS\).}
    \label{table:GramMatrix}
\end{table}
% F is feature map size, Q is projection map size. E.g F = MQ + 1 for RFSF-TRP
% \todo{F = Q?}

\begin{figure}[t]
    \centering
    \includegraphics[width=0.8\linewidth]{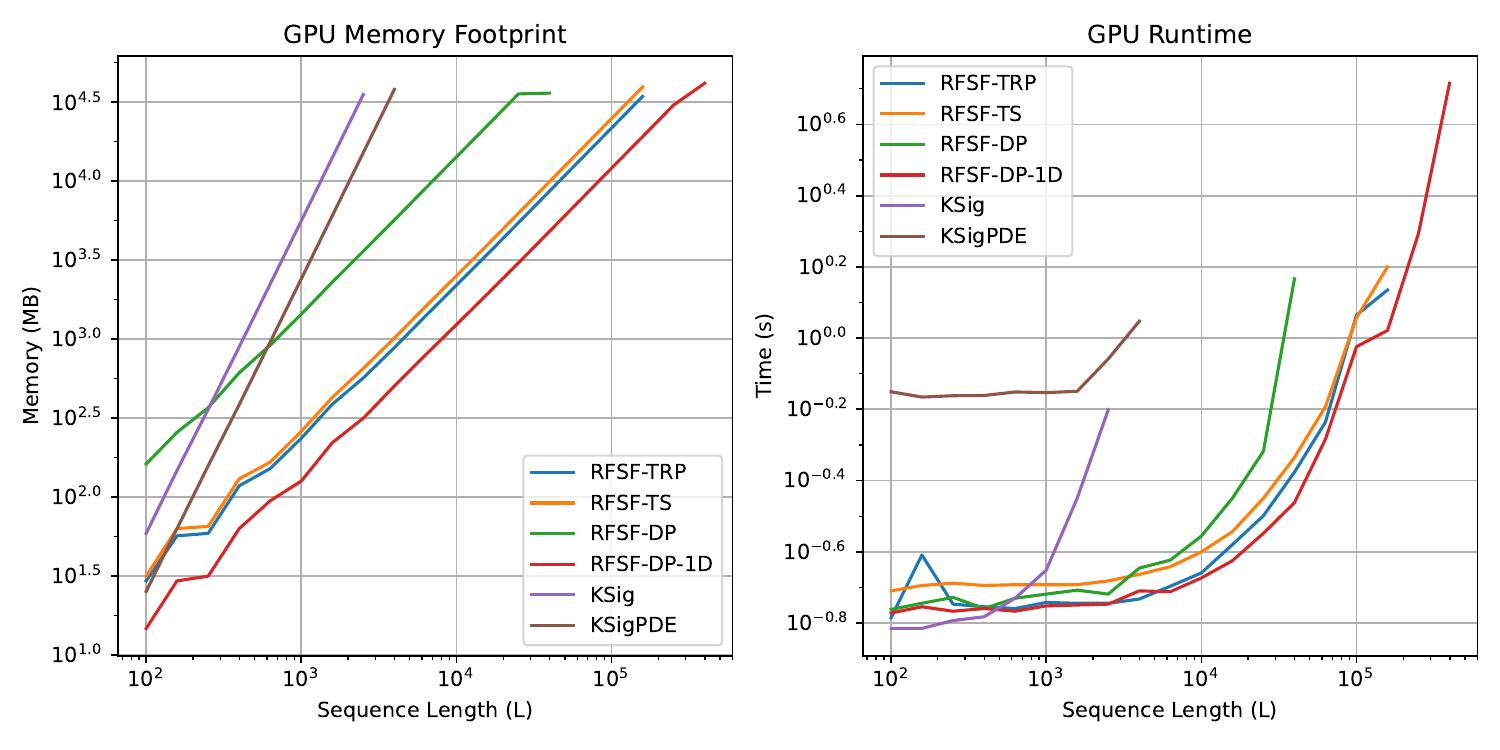}
    \caption{GPU memory footprint and runtime of each algorithm to compute the Gram matrix of 5-dimensional Brownian Motion for varying sequence lengths \(L\). The sequence length was increased for each algorithm until failure due to insufficient memory. We set \(M = 5, D=Q=100, N=10\).}
    \label{fig:Sequence Length Resources}
\end{figure}

This is revealed by the higher gradient of dual methods in the memory footprint plot in Fig. \ref{fig:Sequence Length Resources}.
Due to the increased memory footprint, dual methods can only process sequence lengths \(L \sim 10^3\) before exceeding the memory capacity. Whereas primal methods are able to reach \(L \ge 10^5 \). The runtime behavior is less straightforward due to the nature of GPUs and demonstrates two regimes. In the first regime, there are sufficient GPU cores to parallelize everything leading to a flat growth in runtime. In the second regime, the runtime grows exponentially.
% We are unable to observe the runtime asymptotics due to scaling limitations on GPUs.

% Specified in figure now, N = 10
% \todo{N=1? on figure}

The complexity to obtain feature maps is given in Table \ref{table:Primal Methods Complexity}. All algorithms perform a kernel lift via Random Fourier Features with size \(D\) (Equation \ref{eq:rff}) and differ based on the projection method. These are characterized by their auxiliary projection components and operations. RFSF-TRP is the most expensive due to its use of projection matrices. This incurs a \(MDQ\) storage cost and \(DQ\) operation cost. RFSF-TS is cheaper as it uses hash functions (which scale linearly \(M(D+Q)\)) and applies convolution via FFT which reduces \(Q^2 \to Q \log Q\). The diagonal projection methods have no projection component and perform tensor products locally. These come in two varieties: RFSF-DP uses the 2D-RFF representation (Equation \ref{eq:rff}) like the other algorithms, yielding larger tensors for higher levels, whereas RFSF-DP-1D uses the 1D representation (Equation \ref{eq:rff_1d}) which yields the same size for each level.

\begin{table}[!h]
    \centering
    \begin{tabular}{c|c|c|c|c|c}
        \textbf{Name} & \textbf{Time} & \textbf{Space} & \textbf{Feature} & \textbf{Projection} & \textbf{Output} \\
        \hline
        \textbf{RFSF}   & \(NL D^M\) & \(N L D^M\) & \(MDd\) & - & \(N \times D^M\) \\
        \textbf{RFSF-TRP}  & \(NL MDQ\) & \(NLQ\) & \(MDd\) & \(MDQ\) & \(N \times MQ\)\\
        \textbf{RFSF-TS}  & \(NL M(Q \log Q + Dd)\) & \(NLQ\) & \(MDd\) & \(M(D+Q)\) & \(N \times MQ\) \\
        \textbf{RFSF-DP}  & \(NLD(2^M + Md)\) & \(NL D 2^M\) & \(MDd\) & - & \(N \times 2^M D\) \\
        \textbf{RFSF-DP-1D}  & \(NL MDd\) & \(NLD\) & \(MDd\) & - & \(N \times MD\)
    \end{tabular}
    \caption{Complexity to compute primal signature features. Output is the dimension of the feature map.
    Space is the sum of \textit{Feature} (computation of static features), \textit{Projection} (approximation of outer products) and \textit{Space} (algorithm overhead). The overall space complexity is their sum.
    }
    \label{table:Primal Methods Complexity}
\end{table}

\begin{figure}[t]
    \centering
    \includegraphics[width=0.8\linewidth]{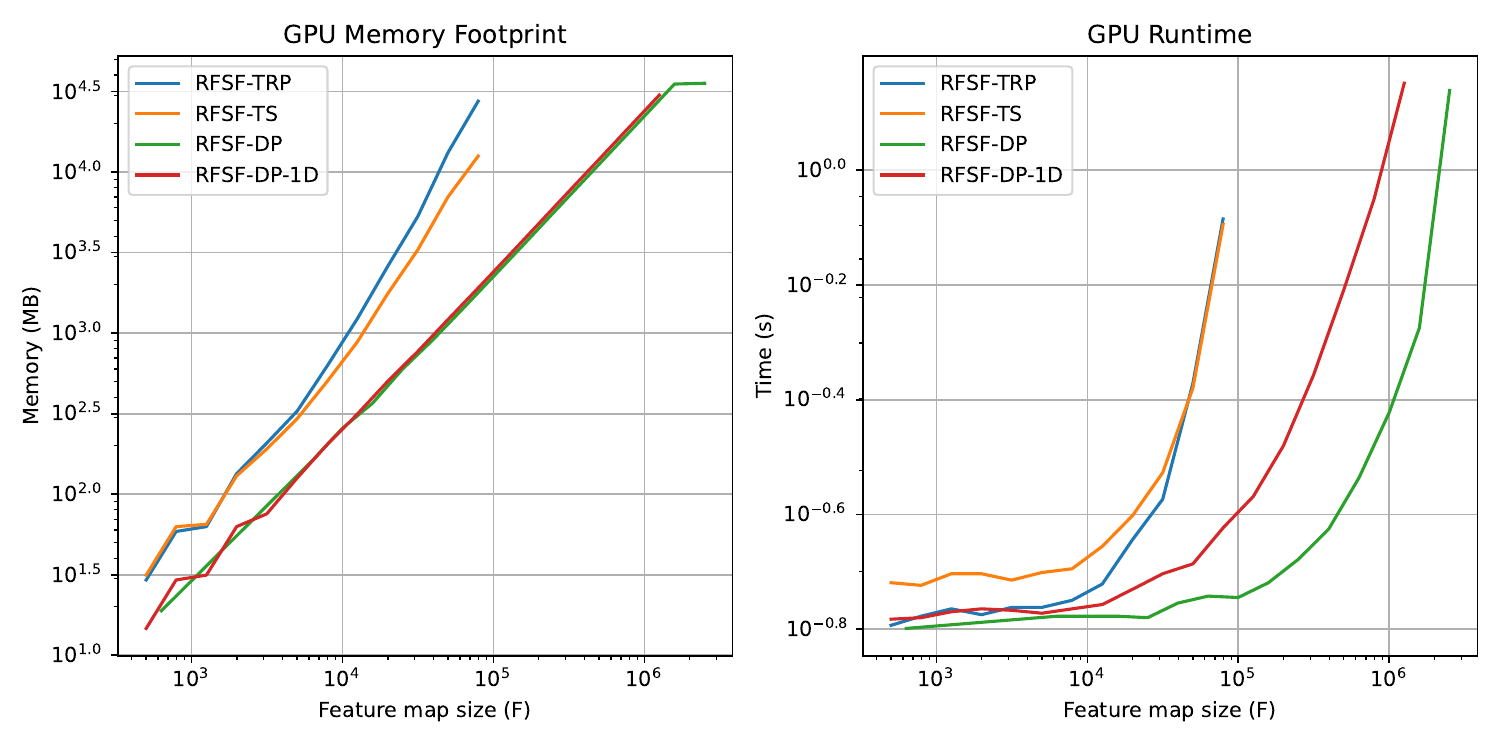}
    \caption{GPU memory footprint and runtime of each algorithm to compute the Gram matrix of 5-dimensional Brownian Motion for increasing feature map size \(F\) until exceeding the memory limit. We set \(M = 5, N = 10, L = 100\). \(D = Q\) is increased to increase \(F\).}
    \label{fig:Feature Map Resources}
\end{figure}

In Fig. \ref{fig:Feature Map Resources}, the computation time and memory footprint of diagonal projection methods (DP and DP-1D) are proportional to their size whereas TRP and TS have an overhead due to their projection components. To tune the feature map size, we look to the next section which studies their accuracies.

% Now pdfs
% \todo[inline]{Fig. \ref{fig:Feature Map Resources} is not vectorized, ie. replace png by pdf. same for other for other figures that are not yet pdf format}

\subsection{Accuracy of Primal Methods}

To measure accuracy, we consider the MAPE between $\sigkernel$ (as it is exact) and the given primal method over distinct pairs in Gram matrix (eq.~\ref{equation:MAPE})
\begin{equation}\label{equation:MAPE}
    \frac{1}{N(N-1)} \sum_{i < j} \left |\frac{\inner{\Phi(\bx^{(i)}, \Phi(\bx^{(j)})}}{\mathrm{\sigkernel}(\bx^{(i)}, \bx^{(j)})} - 1 \right|.
\end{equation}

\begin{figure}[t]
    \centering
    \includegraphics[width=\linewidth]{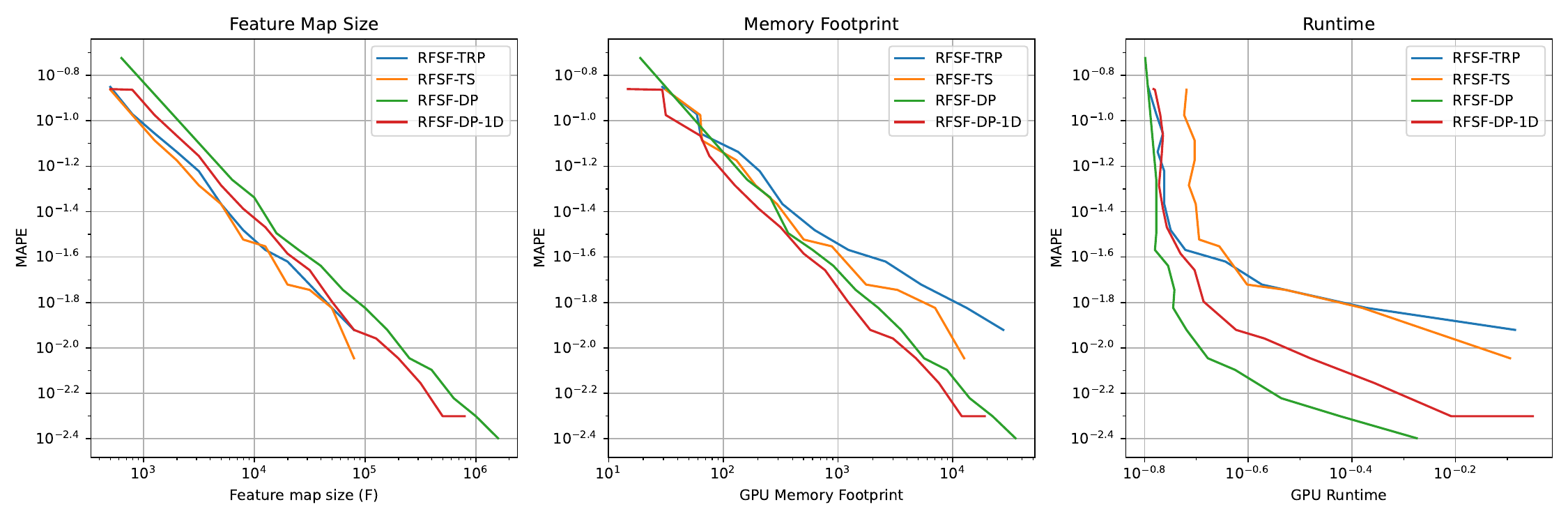}
    \caption{Approximation error of each method against feature map size, memory footprint and runtime. We set \(M = 5, N = 20, L = 100\).}
    \label{fig:Feature Map Accuracy}
\end{figure}

In Fig. \ref{fig:Feature Map Accuracy}, we observe that RFSF-TS and RFSF-TRP offer lower MAPE for a given feature map size \(F\). This may be desirable for downstream applications where \(F\) may have a larger effect on the overall complexity. However, we are unable to obtain higher \(F\) before exceeding the memory capacity. Ultimately, for this task diagonal projection methods allows us to achieve the lowest MAPE. The efficiency is further illustrated by plotting MAPE against memory footprint and runtime.

\begin{figure}[t]
    \centering
    \includegraphics[width=0.8\linewidth]{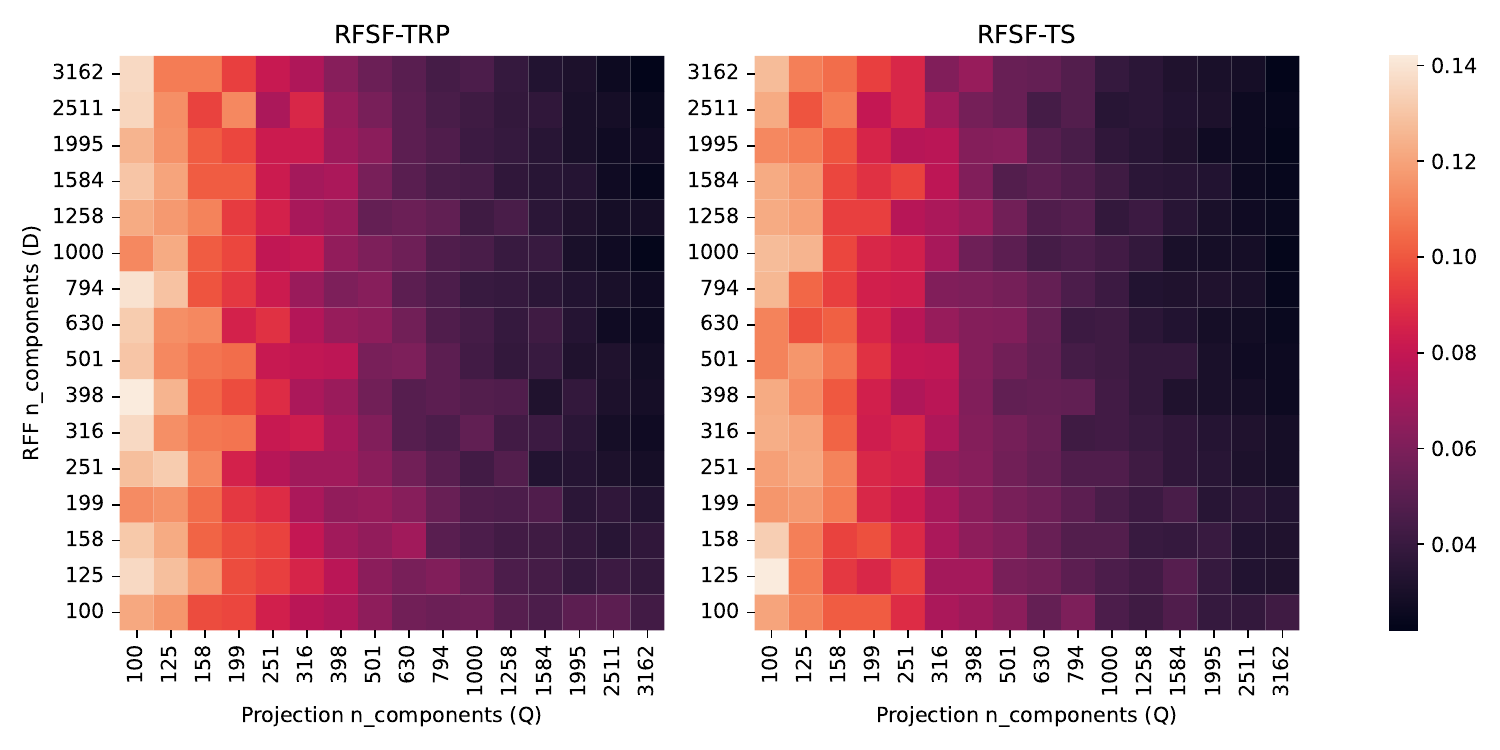}
    \caption{Approximation error of RFSF-TRP and RFSF-TS for various combinations of RFF and projection components, \(D, Q\) respectively. We set \(M = 5, N = 20, L = 100\).}
    \label{fig:Projection RFF MAPE}
\end{figure}

In our experiments, we set \(D = Q\) for RFSF-TRP and RFSF-TS. But potentially one can save resources by using smaller \(D\). Fig. \ref{fig:Projection RFF MAPE} reveals that increasing \(Q\) has a larger effect on decreasing the MAPE.

\subsection{Truncation Level}
\begin{figure}[t]
    \centering
    \includegraphics[width=\linewidth]{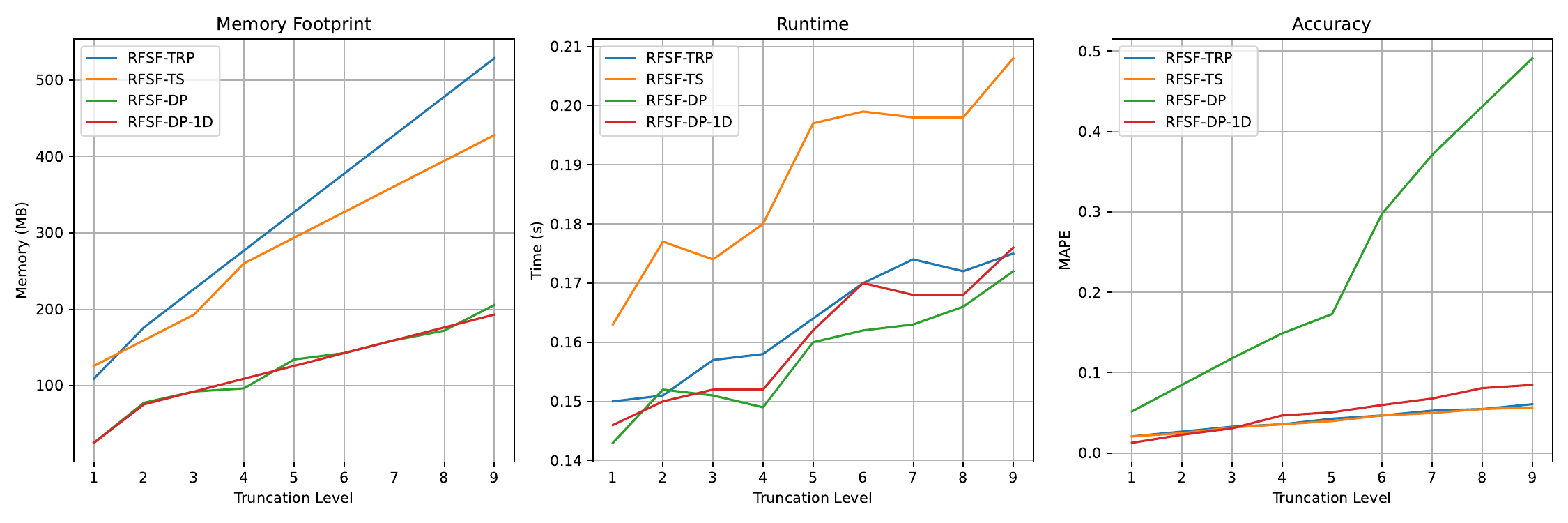}
    \caption{Memory footprint, runtime and accuracy of each method against truncation level \(M\). We set \(N=100, L=100, Q = 1000\). \(D = Q\) except for RFSF-DP which is adjusted to match \(F\).}
    \label{fig:Truncation Level}
\end{figure}

Finally, we consider the scalability of the algorithms with respect to the truncation level \(M\). In order to benefit from higher-order information, high truncation levels are necessary. Fortunately, Tables \ref{table:GramMatrix} and \ref{table:Primal Methods Complexity} indicate all algorithms scale linearly with \(M\) (adjusting RFSF-DP so its feature map size matches the other algorithms \(F = MQ + 1\)). This is numerically verified in Fig. \ref{fig:Truncation Level}. Runtime does not exhibit the behavior observed in Fig. \ref{fig:Sequence Length Resources}. This is because the algorithms compute each level in sequence. Accuracy decreases for fixed \(F\) due to the introduction of higher levels.

\section{Conclusion and Outlook}\label{sec:conclusion}
\texttt{KSig} provides a GPU-accelerated library for efficient computation of various flavours of the signature kernel, bridging theoretical robustness and practical scalability.
We emphasize the following points
\begin{itemize}
    \item \textbf{Modularity and Integration.}
    Following \texttt{Scikit-Learn}'s design, \texttt{KSig} enables seamless integration into diverse kernel learning applications and packages (e.g., supervised learning using SVM, hypothesis testing with MMDs \cite{gretton_kernel_2012, chevyrev_signature_2022}, or unsupervised algorithms such as kernel k-means \cite{dhillon2004kernel}).
    \item \textbf{Choice of Algorithm.} While optimization across possible variations (together with hyperparameter optimization) is recommended for optimal results, the RFSF-DP-1D approach (Section \ref{sec:dp}) offers the most economic computational footprint for given feature size, hence we recommend it as a starting point.
    Its primal formulation (Remark \ref{rem:primal}) avoids Gram matrix computations.
    This allows for linear complexity in the number of sequences and in sequence length without data subsampling!

    \item \textbf{Future Directions.} Future work includes implementing recent advances, such as the forgetting mechanism for time-localized sequence analysis proposed in \cite{toth2024learning} addressing the need to emphasize recent events over distant history in signature kernels, or extending the hypo-elliptic diffusion approach for graphs \cite{toth_capturing_2022} to the kernel learning regime potentially combined with random features for scalability.
\end{itemize}
We encourage community contributions and feedback to further enhance \texttt{KSig}.
The package is available at \texttt{\href{https://github.com/tgcsaba/ksig}{https://github.com/tgcsaba/ksig}.}
\section*{Acknowledgment}
CT was supported by the Mathematical Institute Award by the University of Oxford, DJDC by the EPSRC Centre for Doctoral Training in Mathematics of Random
Systems: Analysis, Modelling and Simulation (EP/S023925/1) and HO by the Hong Kong Innovation and Technology Commission (InnoHK Project CIMDA) and by the EPSRC grant Datasig [EP/S026347/1].
\bibliographystyle{plain}
\bibliography{signatures_survey}

\appendix
\section{Interface of the \texttt{KSig} Package} \label{app:interface}

The hierarchy of the \href{https://github.com/tgcsaba/KSig}{\texttt{KSig}} library is as follows. We outline the relevant files, objects, and their interface. \begin{enumerate*}[label=(\arabic*)] \item \texttt{ksig.kernels} is the most important, which implements various dual and primal time series kernels; \item \texttt{ksig.projections}, \item \texttt{ksig.static.kernels}, and \item \texttt{ksig.static.features} implement options allowing to configure signature kernels with various options; \item \texttt{ksig.preprocessing} implements utilities for preprocessing time series; \item \texttt{ksig.models} implements wrappers on top of SVM models with convenience features. \end{enumerate*} The rest of the files, i.e.~\texttt{ksig.algorithms} and \texttt{ksig.utils} contain under the hood implementations, and are not required to be interacted with by the user. The package uses two main data structures, \texttt{ArrayOnCPU} and \texttt{ArrayonGPU} corresponding to \texttt{NumPy} and \texttt{CuPy} arrays.

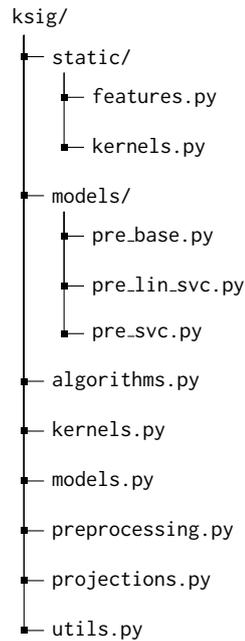
\begin{figure}[h]
\begin{center}
\begin{forest}
for tree={
    font=\ttfamily,
    grow'=0,
    child anchor=west,
    parent anchor=south,
    anchor=west,
    calign=first,
    edge path={
      \noexpand\path [draw, \forestoption{edge}]
      (!u.south west) +(7.5pt,0) |- node[fill,inner sep=1.25pt] {} (.child anchor)\forestoption{edge label};
    },
    before typesetting nodes={
      if n=1
        {insert before={[,phantom]}}
        {}
    },
    fit=band,
    before computing xy={l=15pt},
}
[ksig/
  [static/
    [features.py]
    [kernels.py]
  ]
  [models/
    [pre\_base.py]
    [pre\_lin\_svc.py]
    [pre\_svc.py]
  ]
  [algorithms.py]
  [kernels.py]
  [models.py]
  [preprocessing.py]
  [projections.py]
  [utils.py]
]
\end{forest}
\caption{Hierarchy of submodules in the \texttt{KSig} package} \label{fig:hierarchy}
\end{center}
\vspace{-20pt}
\end{figure}

\subsection{\texttt{ksig.static} folder}
These describe functions (kernels and features), which operate on elements of the time series (hence static as operating in an entry-wise manner on time steps).

\subsubsection{\texttt{ksig.static.kernels}} \label{app:static_kern}

The base class \texttt{Kernel} requires an implementation of the Kernel Gram Matrix \texttt{\_K} which returns a 2D-array with entries \(\kernel(\bx^{(i)}, \by^{(j})\) where \(\by=\bx\) by default. It also requires \texttt{\_Kdiag} which returns a 1D-array with entries \(\kernel(\bx^{(i)}, \bx^{(i)})\). The \texttt{\_\_call\_\_} function calls \texttt{\_K} by default and \texttt{\_Kdiag} if \texttt{diag=True}. Once we instantiate \texttt{K = Kernel(...)}, we can call \texttt{K(X)}, \texttt{K(X, Y)}, \texttt{K(X, diag=True)}.
\newpage

\begin{minted}{python}
from ksig.static.kernels import LinearKernel

K = LinearKernel(...) # Replace with Kernel of choice

X = ... # (Nx, d)
Y = ... # (Ny, d)

# Get Gram Matrix
K(X, X) # (Nx, Nx), equivalent to K(X), K._K(X), K._K(X, X)

# Get Covariance Matrix
K(X, Y) # (Nx, Ny), equivalent to K._K(X, Y)

# Get Gram Matrix diagonal
K(X, diag=True) # (Nx,) equivalent to K._Kdiag(X)
\end{minted}

It is assumed that \(\bx\) is batched \((N, ...)\). The trailing axes represent the kernel domain, either \(L \times d\) for time series or \(d\) for static objects.

There are further base classes with inheritance structure
\begin{center}
\texttt{Kernel <- StaticKernel <- StationaryKernel}.
\end{center}

\texttt{StaticKernel} derives from \texttt{Kernel} and flattens the last two axes as a preprocessing step if \(\bx\) has more than two axes. This is for the setting when \(\bx\) contains time series of shape \(L \times  d\) but we want to treat it as a ``static'' vector of shape \(Ld\).

\begin{minted}{python}
class Kernel(BaseEstimator, metaclass=ABCMeta):
  """Base class for kernels.

  Deriving classes should implement the following methods:
    _K: Computes the kernel matrix between two data arrays.
    _Kdiag: Computes the diagonal kernel entries of a given data array.
    ...
  """
  ...
  @abstractmethod
  def _K(self, X: ArrayOnGPU, Y: Optional[ArrayOnGPU] = None) -> ArrayOnGPU:
    """Computes the kernel matrix.

    Args:
      X: A data array on GPU.
      Y: An optional data array on GPU (if not given uses `X`).

    Returns:
      The kernel matrix between `X` and `Y`, or `X` and `X` if `Y is None`.
    """

  @abstractmethod
  def _Kdiag(self, X: ArrayOnGPU) -> ArrayOnGPU:
    """Computes the diagonal kernel entries.

    Args:
      X: A data array on GPU.

    Returns:
      Diagonal entries of the kernel matrix of `X`.
    """
\end{minted}

The following kernels are implemented. Every kernel defined in this file is static.

\textbf{\texttt{LinearKernel}.} We have the hyperparameter \texttt{scale=1}.
\begin{equation}\label{equation:Linear Kernel}
    \kernel(\bx, \by) = \texttt{scale} \cdot \inner{\bx, \by}
\end{equation}

\begin{minted}{python}
class LinearKernel(StaticKernel):
  """Class for linear static kernel."""

  def __init__(self, scale: float = 1.):
    """Initializes the `LinearKernel` object.

    Args:
      scale: Scaling hyperparameter used to rescale kernel entries.
    """
\end{minted}

\textbf{\texttt{PolynomialKernel}.} We have the hyperparameters \texttt{degree=3, gamma=1, scale=1}.
\begin{equation}\label{equation:Polynomial Kernel}
    \kernel(\bx, \by) = \left(
    \texttt{scale} \cdot \inner{\bx, \by} + \texttt{gamma}
    \right)^{\texttt{degree}}
\end{equation}

\begin{minted}{python}
class PolynomialKernel(StaticKernel):
  """Class for polynomial kernel."""

  def __init__(self, degree: float = 3., gamma: float = 1., scale: float = 1.):
    """Initializes the `PolynomialKernel` object.

    Args:
      degree: Polynomial degree.
      gamma: Offset applied before raising to a power.
      scale: Scaling hyperparameter used to rescale kernel entries.
    """
\end{minted}

\textbf{\texttt{StationaryKernel}} asserts that \(\kernel(\bx, \bx) \equiv 1\) and comes with a \texttt{bandwidth}.
\begin{minted}{python}
class StationaryKernel(StaticKernel):
  """Base class for stationary kernels.

  Stationary kernels considered here have diagonals equal to 1.

  Deriving classes should implement the following methods:
    _K: Computes the kernel matrix between two data arrays.
    ...
  """

  def __init__(self, bandwidth: float = 1.) -> None:
    """Initializes the `StationaryKernel` object.

    Args:
      bandwidth: Bandwidth hyperparameter that inversely scales the input data.
    """
\end{minted}

\textbf{\texttt{RBFKernel}}. We have the hyperparameter \texttt{bandwidth=1}.
\begin{equation}\label{equation:RBF Kernel}
    \kernel(\bx, \by) = \exp\left(
        - \frac{\norm{\bx - \by}^2}{2 \; \texttt{bandwidth}^2}
    \right)
\end{equation}

\begin{minted}{python}
class RBFKernel(StationaryKernel):
  """RBF kernel also called the Gaussian kernel ."""
\end{minted}

\textbf{\texttt{Matern12Kernel}.} We have the hyperparameter \texttt{bandwidth=1}.
\begin{equation}\label{equation:Matern12 Kernel}
    \kernel(\bx, \by) = \exp\left(
        -\texttt{dist}
    \right),\quad
    \texttt{dist} = \frac{\norm{\bx-\by}}{\texttt{bandwidth}}
\end{equation}

\begin{minted}{python}
class Matern12Kernel(StationaryKernel):
  """Matern12 kernel also called the exponential kernel."""
\end{minted}

The Matérn \(3/2\) and \(5/2\) versions are defined analogously for higher orders.

\textbf{\texttt{Matern32Kernel}.} We have the hyperparameter \texttt{bandwidth=1}.
\begin{equation}\label{equation:Matern32 Kernel}
    \kernel(\bx, \by) = (1 + \sqrt{3}\texttt{dist})\exp\left(
        -\sqrt{3}\texttt{dist}
    \right),\quad
    \texttt{dist} = \frac{\norm{\bx-\by}}{\texttt{bandwidth}}
\end{equation}

\begin{minted}{python}
class Matern32Kernel(StationaryKernel):
  """Matern32 (static) kernel ."""
\end{minted}

\textbf{\texttt{Matern52Kernel}.} We have the hyperparameter \texttt{bandwidth=1}.
\begin{equation}\label{equation:Matern52 Kernel}
    \kernel(\bx, \by) = \left(1 + \sqrt{5}\texttt{dist} + \frac{5}{3}\texttt{dist}^2\right)\exp\left(
        -\sqrt{5} \texttt{dist}
    \right),\quad
    \texttt{dist} = \frac{\norm{\bx-\by}}{\texttt{bandwidth}}
\end{equation}

\begin{minted}{python}
class Matern52Kernel(StationaryKernel):
  """Matern52 (static) kernel ."""
\end{minted}

\textbf{\texttt{RationalQuadraticKernel}.} We have the hyperparameters \texttt{bandwidth=1}, \texttt{alpha=1}.
\begin{equation}\label{equation:RationalQuadratic Kernel}
    \kernel(\bx, \by) = \left(
    1 + \frac{\norm{\bx-\by}^2}{2 \; \texttt{alpha} \cdot \texttt{bandwidth}^2}
    \right)^{-\texttt{alpha}}
\end{equation}

\begin{minted}{python}
class RationalQuadraticKernel(StationaryKernel):
  """Rational Quadratic (static) kernel ."""

  def __init__(self, bandwidth: float = 1., alpha: float = 1.):
    """Initializes the rational quadratic kernel.

    Args:
      bandwidth: Bandwidth hyperparameter that inversely scales the input data.
      alpha: Alpha hyperparameter for the rational quadratic kernel.
    """
\end{minted}

\subsubsection{Complexity of static kernels}
All static kernels are \(\bigO{d}\) in time and space, where $d$ is the input dimension.

\begin{table}[t]
    \centering
    \begin{tabular}{c|c|c}
         \textbf{Kernel} & \textbf{Time} & \textbf{Space} \\
         \hline
         \textbf{Linear} & \(d\) & \(d\) \\
         \textbf{Polynomial} & \(d\) & \(d\) \\
         \textbf{RBF} & \(d\) & \(d\) \\
         \textbf{Matérn12} & \(d\) & \(d\) \\
         \textbf{Matérn32} & \(d\) & \(d\) \\
         \textbf{Matérn52} & \(d\) & \(d\) \\
         \textbf{RationalQuadratic} & \(d\) & \(d\)
    \end{tabular}
    \caption{Complexity to compute kernels in \texttt{ksig.static.kernels}.}
    \label{table:Static Kernel Complexity.}
\end{table}

\subsubsection{\texttt{ksig.static.features}} \label{app:static_feat}
The base class \texttt{KernelFeatures} implements a \texttt{fit} and a \texttt{transform} method, which respectively sets up the required variables for the transformation, and sends \(\bx \mapsto \varphi(\bx)\) such that \(\inner{\varphi(\bx), \varphi(\by)} \approx \kernel(\bx, \by)\). Indeed, it is a \texttt{Kernel} itself and sets \(\kernel(\bx, \bx) = \varphi(\bx)^\top \varphi(\by)\), such that $\varphi(\bx), \varphi(\by)$ have dimension $D$, where $D = \texttt{n\_components}$ is a hyperparameter.

Like \texttt{StaticKernel} for \texttt{Kernel}, \texttt{StaticFeatures} defines a subclass of \texttt{KernelFeatures}, such that it flattens the last two axes if there are more than 2 axes.

\begin{minted}{python}
class KernelFeatures(Kernel, TransformerMixin, metaclass=ABCMeta):
  """Base class for featurized kernels.

  Deriving classes should implement the following methods:
    _make_feature_components: Initializes internal variables, called by `fit`.
    _compute_features: Computes the feature map, called by `transform`.
    ...
  """

  def __init__(self, n_components: int = 100,
               random_state: Optional[RandomStateOrSeed] = None):
    """Initializer for the `KernelFeatures` base class.

    Args:
      n_components: Number of feature components to use.
      random_state: A `cupy.random.RandomState`, an `int` seed or `None`.
    """
\end{minted}

Additionally to methods of \texttt{Kernel}, we need to call the \texttt{fit} method, and can call the \texttt{transform} method to obtain the features.

\begin{minted}{python}
from ksig.static.features import RandomFourierFeatures

n_components = 100
KF = RandomFourierFeatures(n_components) # Replace with any KernelFeatures

X = ... # (Nx, d)
Y = ... # (Nx, d)

KF.fit(X) # Assume X, Y are compatible, only need to pass one of them
PhiX = KF.transform(X) # (Nx, n_components)
PhiY = KF.transform(Y) # (Ny, n_components)

KF(X, Y) # (Nx, Ny), Computed using PhiX @ PhiY.T
\end{minted}

The following kernel features are implemented. All classes in this files are static. As the class is static, we can assume that the input \(\bx\) is a vector of dimension \(d\).

\textbf{\texttt{NyströmFeatures}.} \label{docs:NystroemFeatures} Implements the Nyström method from \cite{williams2000using}. Here the \texttt{base\_kernel} \(\kernel\) is set by the user. \(\varphi(\bx)\) has dimension \(\widetilde{D}\), where \(\widetilde{D} \leq D = \texttt{n\_components}\). After sampling \texttt{n\_components} samples during \texttt{fit}, the feature is
\begin{equation}
    \varphi(\bx) = \widetilde{S}^{-1/2} \widetilde{U}_Z^\top \kernel(Z, \bx),\quad
    \kernel(Z, Z) = U_Z S U_Z^{\top}
\end{equation}
where \(\widetilde{U}_Z, \widetilde{S}\) are matrices respectively of dimension \(D \times \widetilde{D}\) and \(\widetilde{D} \times \widetilde{D}\) respectively. It corresponds to truncating the eigenvalues so that \(\widetilde{S}_{ii} > \texttt{EPS} = 1^{-12}\). \(Z\) has \(D = \texttt{n\_components}\) rows which are chosen uniformly without replacement during \texttt{fit} from the fit data.

\begin{minted}{python}
class NystroemFeatures(StaticFeatures):
  """Class for the Nystroem feature map.
     ...
  """

  def __init__(self, base_kernel: Kernel, n_components: int = 100,
               random_state: Optional[RandomStateOrSeed] = None):
    """Initializer for the `NystroemFeatures` class.

    Args:
      base_kernel: The base static kernel object to use.
      n_components: The number of feature components to use in the embedding.
        At most the number of samples of the dataset its fitted on.
      random_state: A `cupy.random.RandomState`, an `int` seed or `None`.
    """
    super().__init__(n_components=n_components, random_state=random_state)
    self.base_kernel = base_kernel
\end{minted}

\textbf{\texttt{RandomFourierFeatures}.} This class implements Random Fourier Features \cite{rahimi2007random} for the RBF Kernel (equation \ref{equation:RBF Kernel}). The class requires a \texttt{bandwidth} parameter corresponding to the RBF kernel. The feature map has dimension $2D$, where $D = \texttt{n\_components}$
\begin{equation}
    \varphi(\bx) = \frac{1}{\sqrt{\texttt{n\_components}}} {\sin(W^\top\bx) \choose \cos(W^\top \bx)},\quad
    W = \frac{Z}{\texttt{bandwidth}}
\end{equation}
where \(Z\) has shape $d \times D$ and i.i.d.~\(\mathcal{N}(0, 1)\) entries. Other stationary kernels are possible, but are not yet implemented. This would correspond to sampling the columns of \(Z\) from the associated spectral measure of another static kernel.

\begin{minted}{python}
class RandomFourierFeatures(StaticFeatures):
  """Class for Random Fourier Features for the Gaussian (RBF) kernel.
     ...
  """

  def __init__(self, bandwidth: float = 1., n_components: int = 100,
               random_state: Optional[RandomStateOrSeed] = None):
    """Initializer for the `RandomFourierFeatures` class.

    Args:
      bandwidth: Bandwidth hyperparameter dividing the input data.
      n_components: The number of feature components to use in the embedding.
      random_state: A `cupy.random.RandomState`, an `int` seed or `None`.
    """
\end{minted}

\textbf{\texttt{RandomFourierFeatures1D}.} Following \cite{gal2015improving}, we can obtain a 1D representation of Random Fourier Features. The class requires a \texttt{bandwidth} hyperparameter corresponding to the RBF Kernel. The feature is of dimension $D$, where $D = \texttt{n\_components}$,
\begin{equation}
    \varphi(\bx) = \sqrt{\frac{2}{\texttt{n\_components}}} \cos(W^\top \bx + \mathbf{b}),\quad
    W = \frac{Z}{\texttt{bandwidth}}
\end{equation}
where \(Z\) has shape $d \times D$ and i.i.d.~\(\mathcal{N}(0, 1)\) entries, \(\mathbf{b}\) has dimension $D$ with i.i.d.~\(\cU[0, 2\pi]\) entries.

\begin{minted}{python}
class RandomFourierFeatures1D(StaticFeatures):
  """Class for Random Fourier Features for the Gaussian (RBF) kernel
     using a 1D representation.
     ...
  """
\end{minted}

\subsubsection{Complexity for Static Features}
We observe that the RFF approach is cheaper than Nyström. However, RFF is only applicable to stationary kernels whereas Nyström applies generally. We denote \(D = \texttt{n\_components}\).

\begin{table}[t]
    \centering
    \begin{tabular}{c|c|c|c|c}
    \multirow{2}{*}{\textbf{Method}} &
      \multicolumn{2}{c}{\textbf{Feature}} &
      \multicolumn{2}{c}{\textbf{Setup}} \\
    & \textbf{Time} & \textbf{Space} & \textbf{Time} & \textbf{Space} \\
    \hline
    \textbf{Nyström} & \(ND^2\) & \(ND\) & \(D^3 + D^2 d\) & \(D^2 + Dd\) \\
    \textbf{RFF} & \(NDd\) & \((N+D)d\) & \(Dd\) & \(Dd\) \\
    \textbf{RFF-1D} & \(NDd\) & \((N+D)d\) & \(Dd\) & \(Dd\)
    \end{tabular}
    \caption{Complexity of feature map methods. \textbf{Feature} describes the cost to compute the features. \textbf{Setup} is the cost to generate the random components.}
    \label{table:features.py Complexity}
\end{table}

%%%%%%%%%%%%%%%%%%%%%%%%%%%%%%%%%%%%%%%%%%%%%%%%%%%%%%%%%%%%%%%%
\subsection{\texttt{ksig.projections}} \label{section:Projections}

\texttt{RandomProjection}s are important to computing low-rank signature kernels which require the computation of tensor products. This class has a \texttt{\_project} method \(\psi\)
% so that \(\inner{\psi(\bx), \psi(\by)} \approx \inner{\bx, \by}\)
and a \texttt{\_project\_outer\_prod} method \(\widehat{\psi}\).
% so that \(\inner{\widehat{\psi}(\bx, \by),  \widehat{\psi}(\bw, \bz)} \approx \inner{\bx \otimes \by, \bw \otimes \bz}\) (possibly more efficienty than using \(\psi(\bx \otimes \by)\).
The output dimension of both maps are controlled by the \texttt{n\_components} attribute. The behavior of these functions depends on the class each projection belongs to and we give further details below.

\begin{minted}{python}
class RandomProjection(BaseEstimator, TransformerMixin, metaclass=ABCMeta):
  """Base class for random projections.

  Random projections are used for computing low-rank signature kernels, where
  a crucial step is computing the outer product of multiple feature arrays.

  Deriving classes should implement the following methods:
    _make_projection_components: Initializes internal vars, called by `fit`.
    _project: Projects an input array, called by `transform`.
    _project_outer_prod: Projects the outer product of a pair of input arrays,
      called by `transform`.
     ...
  """

  def __init__(self, n_components : int = 100,
               random_state: Optional[RandomStateOrSeed] = None):
    """Initializer for `RandomProjection` base class.

    Args:
      n_components: Number of projection components to use.
      random_state: A `cupy.random.RandomState`, an `int` seed or `None`.
    """
\end{minted}

It is crucial that using the \texttt{\_project} $\psi$ and the \texttt{\_project\_outer\_prod} $\widehat{\psi}$ methods are mutually exclusive, and only one can be used within a given \texttt{RandomProjection} instance. The \texttt{fit} method is called to distinguish between the two different behaviors. It must be called respectively using one ($\psi$) or two input arguments ($\widehat{\psi}$) to indicate the desired behavior; see the below code example.

\begin{minted}{python}
from ksig.projections import GaussianRandomProjection
from ksig.utils import outer_prod

n_components = 100
P = GaussianRandomProjection(n_components=n_components) # Replace with RandomProjection of choice

X = ... # (N, dx)
Y = ... # (N, dy)

# Use case 1: Projection of Input
P.fit(X)
P(X) # (Nx, n_components), equivalent to P.transform(X) and P._project(X)

# Use case 2: Projection of Outer Products
P.fit(X, Y)
P(X, Y) # (N, n_components), equivalent to P.transform(X, Y) and P._project_outer_prod(X, Y)

# The above is equivalent to below, but possibly more efficient
XY = outer_prod(X, Y) # (N, dx * dy), batched outer product (XY)_i = X_{i // dx} Y_{i mod dy}

P.fit(XY)
P(XY) # (N, n_components), equivalent to P(X, Y) above
\end{minted}

\subsubsection{Type 1 - Iterative Approximation of Tensor Products} \label{sec:iterative_proj}
First of all, note that these projections are a legacy feature of the library, and instead it is recommended to use the projection classes defined in Section \ref{sec:structured_proj}, which are also discussed in the main text in Section \ref{sec:primalalgs}.
With that noted, using this class of projections corresponds to computing signatures using the double low-rank algorithm  from \cite[Alg.~5]{kiraly_kernels_2019}. This collection of random projections have a \texttt{\_project} method $\bx \mapsto \psi(\bx)$, such that $\inner{\psi(\bx), \psi(\by)} \approx \inner{\bx, \by}$, and set the \texttt{\_project\_outer\_prod} method to \(\widehat{\psi}(\bx, \by) = \psi(\bx \boxtimes \by)\), where $\boxtimes$ corresponds to the Kronecker product, a flattened version of the tensor product.

\textbf{\texttt{GaussianRandomProjection}.} Following \cite{bingham2001random} and \cite{dasgupta2013experiments} the vanilla Gaussian projection is defined as
\begin{equation}
    \psi(\bx) = \frac{Z^\top \bx}{\sqrt{\texttt{n\_components}}}.
\end{equation}
The projection matrix $Z$ is of dimensions $d \times Q$ where $Q = \texttt{n\_components}$ and has i.i.d.~$\mathcal{N}(0, 1)$ entries.

\begin{minted}{python}
class GaussianRandomProjection(RandomProjection):
  """Class for vanilla Gaussian random projections.
     ...
  """
\end{minted}

\textbf{\texttt{SubsamplingProjection}.} A Nyström-type method which subsamples the coordinates of the input
\begin{equation}
    \psi(\bx) = \sqrt{\frac{d}{\texttt{n\_components}}} \bx_{[U]}
    ,
\end{equation}
where \(U\) are indices sampled without replacement from \(\{1, \dots, d\}\). The implementation is optimized to avoid redundancy, i.e.~when calling $(\bx, \by) \mapsto \widehat{\psi}(\bx, \by)$, only the required columns of $\bx \boxtimes \by$ are computed, never needing to compute the full Kronecker product.

\begin{minted}{python}
class SubsamplingProjection(RandomProjection):
  """Class for subsampling as a projection."""
\end{minted}

\textbf{\texttt{VerySparseRandomProjection}.} Similar to \texttt{SubsamplingProjection}, but the indices are selected according to \cite{li2006very}. The sparsity probability \(s(d)\) can either be \texttt{"sqrt"} \(s(d) = 1/\sqrt{d}\) or \texttt{"log"} \(s(d) = \ln d / d\).
\begin{equation}
    \psi(\bx) = \sqrt{\frac{1}{s(d) \cdot \texttt{n\_components}}} Z_{[U, :]}^\top \bx_{[U]},\quad
    Z = R \odot B
    ,
\end{equation}
where \(R, B\) have shape \(d \times \; Q\), where $Q = \texttt{n\_components}$. The entries of \(R\) are i.i.d.~and uniform on \(\pm 1\), i.e. \(\operatorname{Rademacher}(1/2)\), and \(B\) are i.i.d.~\(\operatorname{Bernoulli}(s(d))\). \(U\) corresponds to the rows of \(Z\) which have at least one non-zero entry. This is also optimized so that when calling $(\bx, \by) \mapsto \widehat{\psi}(\bx, \by)$, only the required columns of $\bx \boxtimes \by$ are computed, never needing to compute the full Kronecker product.

\begin{minted}{python}
class VerySparseRandomProjection(RandomProjection):
  """Class for very sparse random projections.
     ...
  """

  def __init__(self, n_components: int = 100, sparsity: str = 'log',
               random_state : Optional[RandomStateOrSeed] = None):
    """Initializer for `VerySparseRandomProjection` class.

    Args:
      n_components: Number of projection components to use.
      sparsity: Rate of sparsity. Possible values are 'sqrt' and 'log'.
      random_state: A `cupy.random.RandomState`, an `int` seed or `None`.
    """
\end{minted}

\subsubsection{Type 2 - Structured Tensor Projections} \label{sec:structured_proj}
Projections in this class compute the projection of the tensor product of multiple vectors in a way that respects the tensorial nature of the input, and correspond to the variations discussed in Section \ref{sec:primalalgs}.
In particular, for a tensor $\bx_1 \otimes \cdots \otimes \bx_m \in (\R^d)^{\otimes m}$, they implement a tensor projection operator $P \in L((\R^d)^{\otimes m}, \R^Q)$, where $Q = \texttt{n\_components}$, such that the algorithm computes $\bx_1 \otimes \cdots \otimes \bx_m \mapsto P(\bx_1 \otimes \cdots \otimes \bx_m)$ without building the input tensor itself. This type also has \texttt{\_project} $\psi$ and \texttt{\_project\_outer\_prod} $\widehat{\psi}$ methods such that the projection of $\bx_1 \otimes \cdots \otimes \bx_m$ is computed recursively as
\begin{align} \label{eq:proj_rec}
    P(\bx_1 \otimes \dots \otimes \bx_m) = \widehat{\psi_m}(\cdots \widehat{\psi_2}(\psi_1(\bx_1), \bx_2), \cdots, \bx_m).
\end{align}

\textbf{\texttt{TensorSketch}.} Following \cite{charikar2002finding} and \cite{pham2013fast}, the \texttt{\_project} function $\psi$ is defined as
\begin{equation}
    \psi(\bx) = \CS(\bx),
\end{equation}
where $\CS$ is defined as in \eqref{eq:cs}. The \texttt{\_project\_outer\_prod} function $\widehat{\psi}$ on the other hand is
\begin{align}
    \widehat{\psi}(\bx, \by) = \FFT^{-1}(\FFT(\bx) \odot \FFT(\CS(\by))),
\end{align}
which assumes that the first input is already sketched. Then, the tensor sketch \eqref{eq:ts_def} is computed via \eqref{eq:proj_rec}, where crucially each projection operator is independent.

\begin{minted}{python}
class TensorSketch(RandomProjection):
  """Class for computing the tensor sketch.
     ...
  """
\end{minted}

\textbf{\texttt{TensorizedRandomProjection}.} Following \cite{sun2021tensor} and \cite{rakhshan2020tensorized}, \texttt{GaussianRandomProjection} can be generalized to the projection of tensorial inputs, see Section \ref{sec:trp}. We retain the same \texttt{\_project} method
\begin{equation}
    \psi(\bx) = \frac{Z^\top \bx}{\sqrt{\texttt{n\_components}}}
    .
\end{equation}
Where \(Z\) has dimension \(d \times Q\) with $Q = \texttt{n\_components}$ and i.i.d.~\(\mathcal{N}(0, 1)\) entries. Then, the \texttt{\_project\_outer\_prod} method is defined as
\begin{align}
    \widehat{\psi}(\bx, \by) = \bx \odot (Z^\top \by),
\end{align}
where $Z$ is as before, and it assumes that $\bx$ has already been projected. Then, the \texttt{TRP} \eqref{eq:trp_def} of a tensor is computed recursively via \eqref{eq:proj_rec}, where crucially each projection operator is sampled independently.

\begin{minted}{python}
class TensorizedRandomProjection(RandomProjection):
  """Class for tensorized random projections.
     ...
  """
  def __init__(self, n_components: int = 100, rank: int = 1,
               random_state : Optional[RandomStateOrSeed] = None):
    """Initializer for `TensorizedRandomProjection` class.

    Args:
      n_components: Number of projection components to use.
      rank: Rank of projection tensors.
      random_state: A `cupy.random.RandomState` or an integer seed or `None`.
    """
\end{minted}

\textbf{\texttt{DiagonalProjection}.} The final projection method is not random. Assume the inputs $\bx, \by$ are of dimension $d \times q$, where $q = \texttt{internal\_size}$. Then, $\psi$ is the identity mapping
\begin{align}
    \psi(\bx) = \bx,
\end{align}
while $\widehat{\psi}$ computes Kronecker products locally along the second dimension, i.e. for $i \in \{1, \dots, d\}$
\begin{align}
    \widehat{\psi}(\bx, \by)_i = \sqrt{d} \cdot \bx_i \boxtimes \by_i.
\end{align}
Note that the approach is only compatible in \texttt{SignatureFeatures} with using as \texttt{static\_features} either variation of RFF, i.e. \texttt{RandomFourierFeatures} or \texttt{RandomFourierFeatures1D}, where $\texttt{internal\_size}$ respectively should be set to $2$ for the former, while for the latter to $1$.

\begin{minted}{python}
class DiagonalProjection(RandomProjection):
  """Class for diagonal projection.
     ...
  """
  def __init__(self, internal_size=2):
    """Initializer for `DiagonalProjection` class."""
    self.internal_size = internal_size
\end{minted}

\subsubsection{Complexity of Structured Tensor Projections}

We do not assume \(d \le Q = \texttt{n\_components}\) as algorithms may have high input dimensionality. Table \ref{table:projections.py Complexity} provides the cost to compute approximations for batch size \(N\) of order \(M\) outer products
\begin{equation}
    \bx_1 \otimes  \cdots \otimes \bx_M
    .
\end{equation}

\begin{table}[!t]
    \centering
    \begin{tabular}{c|c|c|c|c}
    \multirow{2}{*}{\textbf{Method}} &
      \multicolumn{2}{c}{\textbf{Tensor Product}} &
      \multicolumn{2}{c}{\textbf{Feature}} \\
    & \textbf{Time} & \textbf{Space} & \textbf{Time} & \textbf{Space} \\
    \hline
    \textbf{TS} & \(NM Q \log Q\) & \(NQ\) & \(M(Q+d)\) & \(M(Q+d)\) \\
    \textbf{TRP} & \(NM Q\) & \(NQ\) & \(MQd\) & \(MQd\) \\
    \textbf{DPq} & \(Nd q^M\) & \(Nd q^M\) & \(Md\) & \(d\)
    \end{tabular}
    \caption{Complexity of projection methods. \textbf{Outer Product} time is the cost to compute the approximation for an \(M\)-th order tensor product. Space is the size of the approximation. \textbf{Feature} is the time and required to construct the features including the setup cost for the random components.}
    \label{table:projections.py Complexity}
\end{table}

We note that Tensor Sketch (\textbf{TS}) has an \(Q \log Q\) term as the convolution is computed efficiently using the Fast Fourier Transform. Tensorized Random Projection (\textbf{TRP}) has a high memory cost of \(MQd\) due to the multiplication with the Gaussian projection matrices. Diagonal Projection (\textbf{DPq}) has no random components and the setup cost is only the reshape, \(q\) is the \texttt{internal\_size} parameter of the DP algorithm.\textbf{}

%%%%%%%%%%%%%%%%%%%%%%%%%%%%%%%%%%%%%%%%%%%%%%%%%%%%%%%%%%%%%%%%
\subsection{\texttt{ksig.kernels}}

The \texttt{SignatureBase} class defines the options: \begin{enumerate*}[label=(\arabic*)] \item \texttt{n\_levels}: truncation-$M$; \item \texttt{order}: the inclusion map order-$p$; \item \texttt{normalize}: determines whether to return a kernel with diagonals normalized to $1$. \end{enumerate*}
There are two types of normalization: level-wise normalization individually normalizes each signature level independently used by \texttt{KSig} and the primal methods:
\begin{equation}
    \frac{1}{M+1} \sum_{m=0}^{M}
        \frac{\sigkernelm{m}(\bx, \by)}{\sqrt{\sigkernelm{m}(\bx, \bx) \sigkernelm{m}(\by, \by)}}
    .
\end{equation}
Global normalization is used by kernels that do not return signature levels separately, and normalizes the sum of the signature levels; this is used by \texttt{KSigPDE}
\begin{equation}
    \frac{\sigkernelpde(\bx, \by)}{\sqrt{\sigkernelpde(\bx, \bx) \sigkernelpde(\by, \by)}}
    .
\end{equation}

This class is a \texttt{Kernel} and calling instances with input \(\bx, \by\), which have shapes \(N_x \times L_x \times d\) and \(N_y \times L_y \times d\), will return a Gram matrix of shape \(N_x \times N_y\).

\begin{minted}{python}
class SignatureBase(Kernel, metaclass=ABCMeta):
  """Base class for signature kernels.
     ...
  """

  def __init__(self, n_levels: int = 5, order: int = 1,
               difference: bool = True, normalize: bool = True,
               n_features: Optional[int] = None):
    """Initializer for `SignatureBase` base class.

    Args:
      n_levels: Number of signature levels.
      order: Signature embedding order.
      difference: Whether to take increments of lifted sequences in the RKHS.
      normalize: Whether to normalize kernel to unit norm in feature space.
      n_features: Optional, the number of features (state-space dim.).
        Provide this when feeding in flattened sequence arrays of ndim=2.
    """
\end{minted}

The following types of signature kernels are available.

\textbf{\texttt{SignatureKernel}.} We have an additional argument \texttt{static\_kernel} \(\kernel\) of type \texttt{StaticKernel} (see App.~\ref{app:static_kern}). This yields the truncated signature kernel from Section \ref{sec:truncsigkern}.
By default, $\kernel$ is the \texttt{RBFKernel} \eqref{equation:RBF Kernel}. To recover the unlifted signature kernel over $\R^d$, set \texttt{static\_kernel = LinearKernel()}.

\begin{minted}{python}
class SignatureKernel(SignatureBase):
  """Class for full-rank signature kernel."""

  def __init__(self, n_levels: int = 5, order: int = 1,
               difference: bool = True, normalize: bool = True,
               n_features: Optional[int] = None,
               static_kernel: Optional[StaticKernel] = None):
    """Initializes the `SignatureKernel` class.

    Args:
      n_levels: Number of signature levels.
      order: Signature embedding order.
      difference: Whether to take increments of lifted sequences in the RKHS.
      normalize: Whether to normalize kernel to unit norm in feature space.
      n_features: Optional, the number of features (state-space dim.).
        Provide this when feeding in flattened sequence arrays of ndim=2.
      static_kernel: A static kernel from `ksig.static.kernels`.
      ...
    """
\end{minted}

\textbf{\texttt{SignaturePDEKernel}.} We do not have the \texttt{n\_levels} parameter as this approximates the untruncated signature kernel. We can specify \texttt{static\_kernel} for a kernel-lift. This yields the kernel from Section \ref{sec:pdesigkern}.

\begin{minted}{python}
class SignaturePDEKernel(SignatureKernel):
  """Class for full-rank signature-pde kernel."""

  def __init__(self, difference: bool = True, normalize: bool = True,
               n_features: Optional[int] = None,
               static_kernel: Optional[StaticKernel] = None):
    """Initializes the `SignaturePDEKernel` class.

    Args:
      difference: Whether to take increments of lifted sequences in the RKHS.
      normalize: Whether to normalize kernel to unit norm in feature space.
      n_features: Optional, the number of features (state-space dim.).
        Provide this when feeding in flattened sequence arrays of ndim=2.
      static_kernel: A static kernel from `ksig.static.kernels`.
      ...
    """
\end{minted}

\textbf{\texttt{SignatureFeatures}.} This class requires an instantiation of \texttt{StaticFeatures} as the \texttt{static\_features} argument (see App.~\ref{app:static_feat}) and a \texttt{RandomProjection} instance as the \texttt{projection} argument (see App.~\ref{section:Projections}). It inherits from \texttt{KernelFeatures} and the \texttt{transform} method will have shape \(N \times F\) where \(F\) is the signature feature dimension. Concrete values of $F$ for different choices are detailed in Section \ref{sec:primalalgs}. Note that setting \texttt{static\_features = None} will compute vanilla signature features without a static feature lift, while setting \texttt{projection = None}, will compute unprojected signature features, that are tensor-valued.

\begin{minted}{python}
class SignatureFeatures(SignatureBase, KernelFeatures):
  """Class for featurized signature kernels."""

  def __init__(self, n_levels: int = 5, order: int = 1,
               difference: bool = True, normalize: bool = True,
               indep_feat: bool = True, n_features : Optional[int] = None,
               static_features : Optional[StaticFeatures] = None,
               projection: Optional[RandomProjection] = None):
    """Initializes the `SignatureFeatures` class.

    Args:
      n_levels: Number of signature levels.
      order: Signature embedding order.
      difference: Whether to take increments of lifted sequences in the RKHS.
      normalize: Whether to normalize kernel to unit norm in feature space.
      indep_feat: Whether to fit independent features for each tensor product.
      n_features: Optional, the number of features (state-space dim.).
        Provide this when feeding in flattened sequence arrays of ndim=2.
      static_features: A static features object from `ksig.static.features`.
      projection: A projection object from `ksig.projections`.
      ...
    """
\end{minted}

% \textbf{GlobalAlignmentKernel}
% TODO:

\subsection{\texttt{ksig.preprocessing}}
We have two classes to process a collection of sequences. \texttt{SequenceTabulator} ensures sequences are of the same length (with the option to set a max length) and variable time grids by linearly interpolating each time series onto a uniform grid. \texttt{SequenceAugmentor} performs preprocessing options: normalization, lead-lag embedding, adding a time coordinate, and a zero basepoint as the starting step.

Given a collection \texttt{X\_seq} of \(N\) arrays with possibly variable sequence length and dimension \(d\), we provide a pipeline below to generate an array of shape \((N, \widetilde{L}, \widetilde{d})\) ready for our algorithms.
% \(\widetilde{L}, \widetilde{d}\) depend on the chosen options.

\begin{minted}{python}
from ksig.preprocessing import SequenceTabulator, SequenceAugmentor

# Collection (list) of sequences
# Possibly containing nans and differing lengths
X_seq = ...

## Tabulator
tabulator = SequenceTabulator(...)
X_seq = tabulator.fit_transform(X_seq)  # Now an array of shape (N, L, d)

## Augmentor
augmentor = SequenceAugmentor(...)
X_seq = augmentor.fit_transform(X_seq)  # Now an array of shape (N, L_, d_)
\end{minted}

\textbf{\texttt{SequenceTabulator}.} Given a collection \texttt{X\_seq} of \(N\) arrays with shape \((\star, d)\), we can handle missing values and obtain sequences of the same length \(\widetilde{L}\) by interpolation. \texttt{X\_seq} is necessarily a list if the arrays have different lengths, if not \texttt{X\_seq} can be an array of shape \((N, L, d)\) (indicating we want to handle with missing values or reduce the sequence length).

The class has only one parameter: \texttt{max\_len}. By default, \(\widetilde{L}\) is the maximum length of sequences in \texttt{X\_seq}. If \texttt{max\_len} is specified, the length of the output sequences is capped at that value.

\begin{minted}{python}
class SequenceTabulator(BaseEstimator, TransformerMixin):
  """Transformer that tabulates sequences to even length."""

  def __init__(self, max_len: Optional[int] = None):
    """Initializes the `SequenceTabulator` object.

    Args:
      max_len: Maximum length of sequences.
    """
\end{minted}

The interpolation procedure is as follows. Given a sequence \(\set{\bx_0, \dots, \bx_{L-1}}\) of length \(L\), we define a function \(f\) on \([0, 1]\). Set \(f(i/{L-1}) = \bx_i\) and interpolate each dimension separately and linearly.
\begin{equation}
    f(t) = \bx_i + (L-1)\left(t - \frac{i}{L-1}\right)(\bx_{i+1}-\bx_i),\quad
    t \in \left[\frac{i}{L-1}, \frac{i+1}{L-1}\right],\quad
    i \in \set{0, 1, \dots, L-2}
\end{equation}
To obtain the output sequence \(\set{\widetilde{\bx}_0, \dots, \widetilde{\bx}_{\widetilde{L}-1}}\) of length \(\widetilde{L}\) we set \(\widetilde{\bx}_i = f(i/{\widetilde{L}-1})\).

\textbf{\texttt{SequenceAugmentor}.} This class assumes \texttt{X\_seq} is an array of shape \((N, L, d)\). We have a parameter to indicate whether to perform each type of augmentation and auxiliary ones (\texttt{max\_time}, \texttt{max\_len}). The augmentations can be chained and are performed in the following order.
\begin{itemize}
    \item \textbf{\texttt{normalize}.} Ensure all values of \texttt{X\_seq} are in \([-1, 1]\) by dividing by the maximum absolute value found in \texttt{X\_seq}. The output shape is unchanged.
    \item \textbf{\texttt{lead\_lag}.}  Map \(\bx_i = (\bx_i^{(1)}, \dots, \bx_i^{(d)}) \to (\bx_{i+1}^{(1)}, \dots, \bx_{i+1}^{(d)}, \bx_i^{(1)}, \dots, \bx_i^{(d)}) = (\bx_{i+1}, \bx_{i})\). The output shape is \((N, L-1, 2d)\).
    \item \textbf{\texttt{add\_time}.} Obtain a regular grid \(t_i\) on \([0, \texttt{max\_time}]\) of size \(L\) (i.e. \(t_i = \texttt{max\_time} \cdot i / ({L-1})\)), by default \texttt{max\_time} = 1. Map \(\bx_i \to (t_i, \bx_i)\). The output shape is \((N, L, d+1)\).
    \item \textbf{\texttt{basepoint}.} Append \(\mathbf{0} \in \mathbb{R}^d\) to the sequence \(\set{\mathbf{0}, \bx_0, \dots, \bx_{L-1}}\). The output shape is \((N, L+1,d)\).
\end{itemize}

If \texttt{max\_len} is specified and the length is greater than \texttt{max\_len}, we downsample by interpolation.

\begin{minted}{python}
class SequenceAugmentor(BaseEstimator, TransformerMixin):
  """Transformer that tabulates sequences to even length."""

  def __init__(self, add_time: bool = True, lead_lag: bool = True,
               basepoint: bool = True, normalize: bool = True,
               max_time: float = 1., max_len: Optional[int] = None):
    """Initializes the `SequenceAugmentor` object.

    Args:
      add_time: Whether to augment with time coordinate.
      lead_lag: Whether to augment with lead-lag.
      basepoint: Whether to augment with basepoint.
      normalize: Whether to normalize time series.
      max_time: Maximum time if `add_time is True`.
      max_len: Maximum length of sequences.
    """
\end{minted}

\subsection{\texttt{ksig.models}}

This folder contains classes to generate Support Vector Classifiers (handling multiclass via a one-vs-one scheme) which have built-in hyperparameter tuning (via stratified cross validation and grid search). Computing the SVC is made more efficient by precomputing certain quantities (increased memory for reduced runtime) and reusing it across the grid search. Furthermore, we have two versions of SVC. \texttt{PrecomputedKernelSVC} performs standard dual SVC, leading to cubic \(N^3\) complexity in the number of sequences \(N\). \texttt{PrecomputedFeatureLinSVC} leverages a feature map approximation of the kernel leading to a complexity \(NF^2\) which is linear in \(N\). \(F\) is the dimension of the feature map.

Each class lives in its own file. \texttt{PrecomputedSVCBase} in \texttt{pre\_base.py}, \texttt{PrecomputedFeatureLinSVC} in \texttt{pre\_lin\_svc.py} and \texttt{PrecomputedKernelSVC} in \texttt{pre\_svc.py}.

\textbf{\texttt{PrecomputedSVCBase}.} This base class performs the pipeline: precomputing quantities, hyperparameter tuning and prediction. It requires deriving classes to define how to perform SVC (dual or primal) and specify what needs to be precomputed (either the kernel matrix or feature matrix). The interface is simply

\begin{minted}{python}
# Covariates - Shape (N, L, d)
X_train = ...
X_test = ...

# Labels - Shape (N,)
y_train = ...

svc_model = ...
# Perform hyperparameter tuning if svc_grid specified
svc_model.fit(X_train, y_train)
# Predict
y_pred = svc_model.predict(X_test)
\end{minted}

The parameters \texttt{svc\_grid}, \texttt{cv} and \texttt{n\_jobs} pertain to hyperparameter tuning, e.g. the regularization parameter. \texttt{svc\_hparams} are additional fixed parameters that can be specified, e.g. error tolerance.

\begin{minted}{python}
class PrecomputedSVCBase(BaseEstimator, ClassifierMixin, metaclass=ABCMeta):
  """Base class for precomputed SVC models.

  Deriving classes should implement the following methods:
    _get_svc_model: Constructs and returns the SVC model to use.
    _precompute_model_inputs: Precomputes the inputs fed to the SVC model.

  Warning: This class should not be used directly, only derived classes.
  """
  def __init__(self,
               kernel: Callable,
               svc_hparams: Dict = {},
               svc_grid: Optional[Dict] = None,
               cv: int = 3,
               n_jobs: int = -1,
               need_kernel_fit: bool = False,
               has_transform: bool = False,
               batch_size: Optional[int] = None):
    """Initializer for `PrecomputedSVCBase`.

    Args:
      kernel: A callable for computing the kernel or feature matrix.
      svc_grid: Hyperparameter grid to cross-validate over, set to `None`
        for no cross-validation.
      svc_hparams: Additional hyperparameters for `SVC`.
      cv: Number of CV splits.
      n_jobs: Number of jobs for `GridSearchCV`.
      need_kernel_fit: Whether the kernel needs to be fitted to the data.
      has_transform: Whether the kernel has a feature transform.
      batch_size: If given, compute the kernel matrix in chunks of shape
        `[batch_size, batch_size]`, or in chunks of shape [batch_size, ...]
        if `has_transform` is set to true.
    """
\end{minted}

\textbf{\texttt{PrecomputedFeatureLinSVC}.} Same parameters as the base class but the kernel needs to be able to generate features. I.e. it is of type \texttt{ksig.static.features.KernelFeatures}. Uses primal SVC to compute solution with complexity \(NF^2\) where \(N\) is number of samples and \(F\) is the size of the feature map approximation to kernel. Note that setting \texttt{on\_gpu = True} requires the \texttt{CUML} package to be installed.

\begin{minted}{python}
class PrecomputedFeatureLinSVC(PrecomputedSVCBase):
  """`LinearSVC` with precomputed features with optional cross-validation."""

  def __init__(self,
               kernel: KernelFeatures,
               svc_hparams: Dict = {},
               svc_grid: Optional[Dict] = None,
               cv: int = 5,
               n_jobs: int = -1,
               need_kernel_fit: bool = False,
               batch_size: Optional[int] = None,
               on_gpu: bool = False):
    """Initializer for `PrecomputedFeatureLinSVC`.

    Args:
      kernel: A callable for computing features.
      svc_grid: Hyperparameter grid to cross-validate over, set to `None`
        for no cross-validation.
      svc_hparams: Additional hyperparameters for `LinearSVC`.
      cv: Number of CV splits.
      n_jobs: Number of jobs for `GridSearchCV`.
      need_kernel_fit: Whether the features need to be fitted to the data.
      batch_size: If given, compute the feature matrix in chunks of shape
        `[batch_size, ...]` in order to save memory.
      on_gpu: Whether to use the GPU implementation or not.
    """
\end{minted}

\textbf{\texttt{PrecomputedKernelSVC}.} Same parameters as the base class. Uses dual SVC to compute solution with cubic complexity \(N^3\) where \(N\) is number of samples.

\begin{minted}{python}
class PrecomputedKernelSVC(PrecomputedSVCBase):
  """`SVC` with precomputed kernel with optional `GridSearchCV`."""
  def __init__(self,
               kernel: Kernel,
               svc_hparams: Dict = {},
               svc_grid: Optional[Dict] = None,
               cv: int = 3,
               n_jobs: int = -1,
               need_kernel_fit: bool = False,
               batch_size: Optional[int] = None,
               has_transform: bool = False):
    """Initializer for `PrecomputedKernelSVC`.

    Args:
      kernel: A callable for computing the kernel matrix.
      svc_grid: Hyperparameter grid to cross-validate over, set to `None`
        for no cross-validation.
      svc_hparams: Additional hyperparameters for `SVC`.
      cv: Number of CV splits.
      n_jobs: Number of jobs for `GridSearchCV`.
      need_kernel_fit: Whether the kernel needs to be fitted to the data.
      batch_size: If given, compute the kernel matrix in chunks of shape
        `[batch_size, batch_size]`, or in chunks of shape [batch_size, ...]
        if `has_transform` is set to true.
      has_transform: Whether the kernel has a feature transform.
    """
\end{minted}

% \subsubsection{Example}\label{section:ksig.models example}
% We refer to the example given in section \ref{sec:primalimpl}. Here we tune the regularisation parameter.
% \inputminted{python}{code/main/primal_experiment.py}

% \textbf{RandomWarpingSeries}
% TODO:
\end{document}